\documentclass[11pt]{article}   

\usepackage[left=2.54cm, right=2.54cm, top=2.54cm]{geometry}
\usepackage{appendix}

\usepackage{graphicx}
\usepackage{amsmath,amsfonts,bm} 
\usepackage{algorithm}
\usepackage{algpseudocode}
\usepackage{multirow} 
\usepackage[makeroom]{cancel} 

\emergencystretch=\maxdimen
\providecommand{\keywords}[1]
{
	\small	
	\textbf{\textit{Keywords:}} #1
}

\usepackage[font=small,labelfont=bf]{caption}

\newcommand{\reals}{ {\mathbb R} }
\newcommand{\naturalnums}{ {\mathbb N} }

\newcommand{\ldsmodel}[1]{\mathcal{M}^{(#1)}}

\newcommand{\bx}{{\bf x}}
\newcommand{\bu}{{\bf u}}
\newcommand{\bX}{{\bf X}}
\newcommand{\bU}{{\bf U}}
\newcommand{\bY}{{\bf Y}}

\newcommand{\bt}[1]{\widetilde{\mathbf #1}} 

\newcommand{\X}{\mathcal{X}}

\renewcommand{\S}{\mathcal{S}}
\newcommand{\A}{\mathcal{A}}
\newcommand{\Q}{\mathcal{Q}}
\newcommand{\R}{\mathcal{R}}
\newcommand{\D}{\mathcal{D}}
\newcommand{\Tr}{\mathcal{T}}

\newcommand{\T}{\mathrm{T}}

\DeclareMathOperator*{\argmax}{argmax}

\usepackage{gensymb} 

\usepackage{appendix} 

\begin{document}
	\title{RLOC: Neurobiologically Inspired Hierarchical Reinforcement Learning Algorithm for Continuous Control of Nonlinear Dynamical Systems}
	\author{Ekaterina Abramova\textsuperscript{$\ast$},
		Luke Dickens \textsuperscript{$\ast$},
		Daniel Kuhn \textsuperscript{$\ast$},
		Aldo Faisal\textsuperscript{$\ast$,$\dagger$,$\ddagger$}}
	\date{}
	\maketitle
	\begin{center}
		$\ast$ Department of Computing, Imperial College London, London, UK
		\\
		$\dagger$ Department of Bioengineering, Imperial College London, London, UK
		\\
		$\ddagger$ MRC Clinical Science Center, London, UK
		\\
		\bigskip
	\end{center}
	
	\begin{abstract}
		Nonlinear optimal control problems are often solved with numerical methods that require knowledge of system's dynamics which may be difficult to infer, and that carry a large computational cost associated with iterative calculations. We present a novel neurobiologically inspired hierarchical learning framework, Reinforcement Learning Optimal Control, which operates on two levels of abstraction and utilises a reduced number of controllers to solve nonlinear systems with unknown dynamics in continuous state and action spaces. Our approach is inspired by research at two levels of abstraction: first, at the level of limb coordination human behaviour is explained by linear optimal feedback control theory. Second, in cognitive tasks involving learning symbolic level action selection, humans learn such problems using model-free and model-based reinforcement learning algorithms. We propose that combining these two levels of abstraction leads to a fast global solution of nonlinear control problems using reduced number of controllers. Our framework learns the local task dynamics from naive experience and forms locally optimal infinite horizon Linear Quadratic Regulators which produce continuous low-level control. A top-level reinforcement learner uses the controllers as actions and learns how to best combine them in state space while maximising a long-term reward. A single optimal control objective function drives high-level symbolic learning by providing training signals on desirability of each selected controller. We show that a small number of locally optimal linear controllers are able to solve global nonlinear control problems with unknown dynamics when combined with a reinforcement learner in this hierarchical framework. Our algorithm competes in terms of computational cost and solution quality with sophisticated control algorithms and we illustrate this with solutions to benchmark problems.
	\end{abstract}
	
	
	\section{Introduction} \label{sec:intro} 
	Many real-world control problems involve complex, nonlinear and often unknown dynamics, such as those encountered in robotics, manufacturing, prosthetics, piloting/steering and other engineering areas.
	Analytical solutions, based on the Hamilton-Jacobi-Bellman partial differential equation, are often limited to stylised problems, while numerical solutions tend to depend on partial/full knowledge of the system's dynamics in order to form approximate solutions.
	A central issue to finding approximate optimal control is the speed of learning and data efficiency, the number of trials required to learn a task. 
	Despite recent advances in combining deep learning for feature representations with reinforcement learning (RL) for choosing actions that maximise cumulative future reward, resulting in deep reinforcement learning \cite{Mnih2015,Silver2016,Silver2017b,Levine2015}, the problem of data inefficiency remains an issue.
	The resulting high computational cost means that the optimal control of nonlinear dynamical systems presents a fundamental challenge, especially if the system to be controlled is not well characterised \cite{kappen2011b}. 
	
	Our contribution to the nonlinear control community is a proof-of-principle hierarchical control framework which is able to quickly learn continuous control of nonlinear dynamical systems with unknown dynamics for infinite-horizon problems.
	The proposed algorithm, Reinforcement Learning Optimal Control (RLOC), is a hierarchical two-level closed feedback loop control framework, which assembles a global control policy through primitive fragments.
	A high-level reinforcement learning decision maker (agent) learns how to best combine low-level continuous infinite horizon linear quadratic regulator controllers (LQR$_{\infty}$) for solving nonlinear global control tasks with unknown dynamics.
	A distinguishing feature of the architecture is a single learning signal, the low-level control cost function, which drives the high-level agent's learning, thus feeding the low-level error signals back into the system forming a closed loop control system.
	This paper focuses on designing control laws for deterministic nonlinear systems with unknown dynamics, in continuous time, state and action spaces. 
	
	\textbf{Neurobiological Motivation}
	The two levels of RLOC's hierarchy are neurobiologically motivated by research which indicates that the brain learns and executes complex nonlinear control tasks, such as object manipulation and sensorimotor control, using computations on the two levels of abstraction.
	First, at the level of actuation, it has been shown that the human brain controls the redundant degrees of freedom of reaching movements in a manner well predicted by linear Optimal Feedback Control theory \cite{scott2004optimal}. 
	The evidence strongly suggests that the brain minimises motor noise \cite{faisal2008noise} and/or effort \cite{o2009dissociating}, which are assumed to be quadratic cost functions of the muscle control signals \cite{scott2004optimal,todorov2004optimality}. 
	Moreover, the control of actuators appears to be represented in the form of local controllers of the state space, and characteristic linear control "motor primitives" were found to operate at the actuator level \cite{mussa1994linear,d2003combinations}.
	It is therefore plausible for low-level continuous RLOC controllers to utilise simple linear controllers at the low-level of actuator control and we propose to use linear quadratic regulators in an infinite control horizon setting, which are optimal with respect to its cost function and the learnt dynamics.
	Cognitive neuroscience experiments involving computational models and cognitive neuroimaging, have recently shown that flexible closed-loop control is performed without tracking a pre-defined trajectory \cite{Hellyer2014}.
	This is in line with our framework which uses a closed-loop feedback system with low-level control quadratic costs driving high-level learning.
	Second, at the level of optimal action selection in symbolic serial decision making tasks, the results show that humans, monkeys and mice produce behaviour and brain signals consistent with known reinforcement learning algorithms.
	Evidence for a reward prediction error, which calculates the difference between actual and expected reward of a chosen action in a certain state, has been shown to be encoded in the activity of the mid-brain dopaminergic neurons of the substantia nigra and ventral tegmental area, that form a part of the basal ganglia circuit \cite{schultz1997neural,d2008bold}. 
	In the context of reinforcement learning this reward prediction error signal is used to learn values for action choices that maximise an expected future reward and is known as 'model-free' reinforcement learning \cite{sutton1998reinforcement}.
	In contrast, another class of reinforcement learning builds internal models of how actions and states interact, enabling simulating an experience. 
	This 'model-based' reinforcement learning calculates state prediction error, which measures the surprise in the new state given the current estimate of the state-action-state transition probabilities and a neural signature for this error is present in the intraparietal sulcus and lateral prefrontal cortex \cite{glascher2010states}.
	The above research suggests symbolic task representation in the brain and thus we choose to employ a model-free reinforcement learning agent as a high-level symbolic decision maker which learns a globally optimal control policy by linking low-level continuous controllers to be activated by the agent in areas of state space. 
	Even though we employ a hierarchical structure for the control algorithm, since hierarchy is undoubtedly present in the anatomy, it is unclear whether the biological neural controllers implement their strategies in a hierarchical way, thus we note that this is an engineering construct with an assumed biological implementation.

	\subsection{Related Work}
	Control of complex, nonlinear dynamical systems is a vast research topic, with state-of-the-art solutions broadly separated into model-based, model-free and hybrid methods, the most prominent of which are compared and contrasted to RLOC below.\\

	\textbf{Model-based Methods}
	One of the most studied control strategies in engineering is the model-based control, where the plant dynamics are made explicitly known to the algorithm a priori, the most prominent algorithms are: nonlinear model predictive control \cite{grune2011nonlinear}, approximate dynamic programming \cite{atkeson2008random}, motion planning \cite{Tedrake2009b} and differential dynamic programming \cite{theodorou2010stochastic}.
	The iterative linear quadratic regulator (iLQR) control algorithm \cite{li2004iterative} uses iterative linearisation of the known nonlinear dynamics around candidate trajectories that are improved recursively.
	This algorithm is of particular relevance to RLOC since we are able to show that a reduced number of equivalent linear quadratic regulators can be efficiently combined by a high-level reinforcement learning agent for global control, which alleviates the curse of dimensionality, a problem becoming intractable as the number of states grows \cite{Bellman1958}.
	
	Another important class of algorithms which is of relevance to the robotics community relies on probabilistic representations over motor commands. 
	The basis of probabilistic control models stems from work on utilising Bayesian learning.
	For example, Demiris et.al. \cite{Dearden2005a} used Bayesian networks to learn and thus represent forward models capable of predicting effects of robot actions on its motor system and the environment. 
	Current state-of-the-art learning control algorithm in terms of data efficiency is known as Probabilistic Inference for Learning Control (PILCO) \cite{deisenroth2011pilco}.
	It combines Bayesian and function approximation approaches, and uses Gaussian Processes to model and express uncertainty of the unknown dynamics and policy, while simultaneously learning both. 
	It is a highly data efficient algorithm, which requires little interaction time with the physical system, yet suffers from the need for intensive off-line parameter optimisation calculations \cite{rasmussen2008probabilistic}.
	In a more recent work \cite{Kamthe2017} a moment matching approach based on probabilistic model predictive control, GP-MPC, has been developed, which improves on PILCO's need to look over the full planning horizon and to optimise hundreds/thousands of parameters per control dimension. 
	The algorithm is reported to halve PILCO's physical system interaction time, and to improve on the off-line calculations time, however the exact time requirements are not reported.
	We show that RLOC's simple representation is able to learn on-line an optimal global control policy within a matter of minutes in order to control the cart-pole from all starting positions in state space. \\

	\textbf{Model-free Methods}
	Real world tasks are often not well characterised, are non-stationary and/or difficult to infer, thus requiring the use of adaptive machine learning methods capable of working with unknown task dynamics. 
	One such prominent technique maps standard reinforcement learning algorithms \cite{sutton1998reinforcement} to continuous state, time and action spaces \cite{doya2000reinforcement}.
	However this is computationally expensive due to the need for state-action space exploration when learning the global control policy.
	The contribution of our algorithm is a reduction in this computational requirement, achieved via a simplification of the overall state complexity due to hierarchical structure and the use of continuous low-level controllers which leads to a large reduction in computational requirements.  
	
	While many have attempted to address the curse of dimensionality \cite{singh1992transfer, dayan1993feudal, dean1995decomposition}, a class of algorithms known as Hierarchical Reinforcement Learning \cite{mcgovern1998hierarchical,barto2003recent} alleviates this by employing temporal abstractions known as "options", which are actions resulting in state transitions of variable durations \cite{Sutton2015a}.
	The architecture typically involves manually decomposing a task into a set of sub-tasks and learning both high-level and low-level policies, which overall converge to a recursively optimal policy \cite{mcgovern1998hierarchical}, as for example is used in robotic tasks \cite{lee2006quadruped} and a general purpose hierarchical algorithm MAXQ \cite{dietterich2000hierarchical}. 
	More recently, a deep hierarchical reinforcement learning framework based on symbolic policy sketches \cite{Andreas2016}, was able to learn deep multitask policies with minimal supervision for tasks with delayed and/or long-term reward structure.
	Although this class of algorithms are similar to RLOC in terms of the hierarchical structure, we note the following differences: (a) each RLOC high-level action (option equivalent) activates a low-level LQR controller, responsible for continuous actuator control. 
	This means that the high-level RLOC policy learns which low-level actions are best suited for each area of state space, rather than optimising a pre-defined sub-goal's policy, (b) a single LQR cost function drives the whole algorithm's learning.    
	A framework of using low-level actions has been utilised in a hierarchical probabilistic generative model developed by Storkey et al. \cite{williams2008modelling}, which uses a superposition of sparsely activated "motion primitives", a special kind of parametrised policies exploiting the analogy between autonomous differential equations and control policies \cite{ijspeert2003learning}. 
	This work was extended to using many motion primitives in a hierarchical reinforcement learning framework, "dynamic movement primitives", allowing to learn arbitrary parametrisation of the abstracted level rather than fixing it \emph{a priori}, forming building blocks of motion ("parametrized options") in continuous time \cite{neumann2009learning}. 
	Recently Schaal et al., proposed a probabilistic reinforcement learning method, Policy Improvement with Path Integrals (PI$^2$), which operated on a 12-D system \cite{theodorou2010reinforcement}.
	New skill discovery \cite{konidaris2009skill} method has also been proposed that establishes, on-line, chains of options in continuous domains.\\

	\textbf{Hybrid Methods}
	Another class of models combines model-free and model-based algorithms, known as hybrid models. 
	These have been applied to learning reward functions from demonstration \cite{atkeson1997robot}, and to obtaining approximate models of the dynamics used in differential dynamic programming and policy gradient evaluations, which required only a small number of real-life trials \cite{abbeel2006using}.
	More recently Abbeel et al. \cite{Levine2014a}, designed an algorithm which learnt stationary nonlinear policies for solutions to unknown dynamical systems using priors and an improved iLQG algorithm. 
	As compared to hybrid models, RLOC offers an alternative framework which learns linear and nonlinear plant dynamics on-line and does not rely on approximate models.
	We show using two low-dimensional standard benchmark problems that RLOC outperforms/matches the benchmark algorithms: naive switching, LQR, state-of-the-art iLQR and state-of-the-art PILCO, in terms of accuracy and computational requirements, thus forming a promising framework for addressing this important class on nonlinear control problems with unknown dynamics.\\

	\textbf{Frameworks Similar to RLOC}
	A number of frameworks combing reinforcement learning with linear quadratic regulators (LQR) exist in the literature.
	Current literature combining RL with LQR relies on full knowledge of the system dynamics and expertly built fine-tuned controllers, thus not truly learning control problems from the bottom-up.
	The aim of our RLOC research is to build upon the current algorithms in this area, present improvements in their design, show that we can solve standard benchmark problems with unknown dynamics and provide faster control solutions.
	We give a brief overview of the present-day algorithms in the literature which possess similarity to RLOC and outline our contribution.
	\begin{enumerate}
		\item 
		Radlov et al. \cite{randlov2000combining} combined a single hand-crafted LQR controller with a reinforcement learning agent in order to solve a double link pendulum task containing a motor at each joint.
		The controller, however, was only activated once the agent drove the system to a pre-defined state, positioned near the target, hence pre-fixing the switching condition in advance.
		The LQR controller was expertly located in this area which guaranteed stability, allowing the system to learn faster when compared to using the agent alone. 
		\item 
		The above approach was extended by Yoshimoto et al. \cite{yoshimoto2005acrobot}, who used a reinforcement learning agent to learn how to best combine controllers previously devised for controlling the double link pendulum task with known dynamics in different areas of state space for global use.
		The authors showed that a rather large number of learning trials (when compared to RLOC) was required to train the system.
		The final policy was only able to control the double link pendulum to the upright target from a stable equilibrium starting state, while RLOC is able to control the system from any starting state.
	\end{enumerate}

	We advance the above approaches using our algorithm, RLOC, which in contrast to previous work is able to:
	(a) learn the system dynamics from experience on-line, thus not relying on explicitly pre-defined dynamics; 
	(b) learn low-level LQR controllers on-line, which are not expertly built or pre-defined based on specific task requirements;
	(c) learn a global control policy for controlling a given system from any area of state space to a given target, as opposed to single start-end control; 
	(d) utilise a reduced number of episodes in order to train equivalent systems, demonstrating a much faster learning rate than state-of-the-art, as tested on the same machine for the same benchmark problems.
	
	We proceed with establishing the formal structure of RLOC algorithm, where low-level control, local linear approximations and high-level control are each presented in detail.

    \section{Methods}
\begin{figure}[!ht]
    	\centering
    	\includegraphics[width=0.75\textwidth]{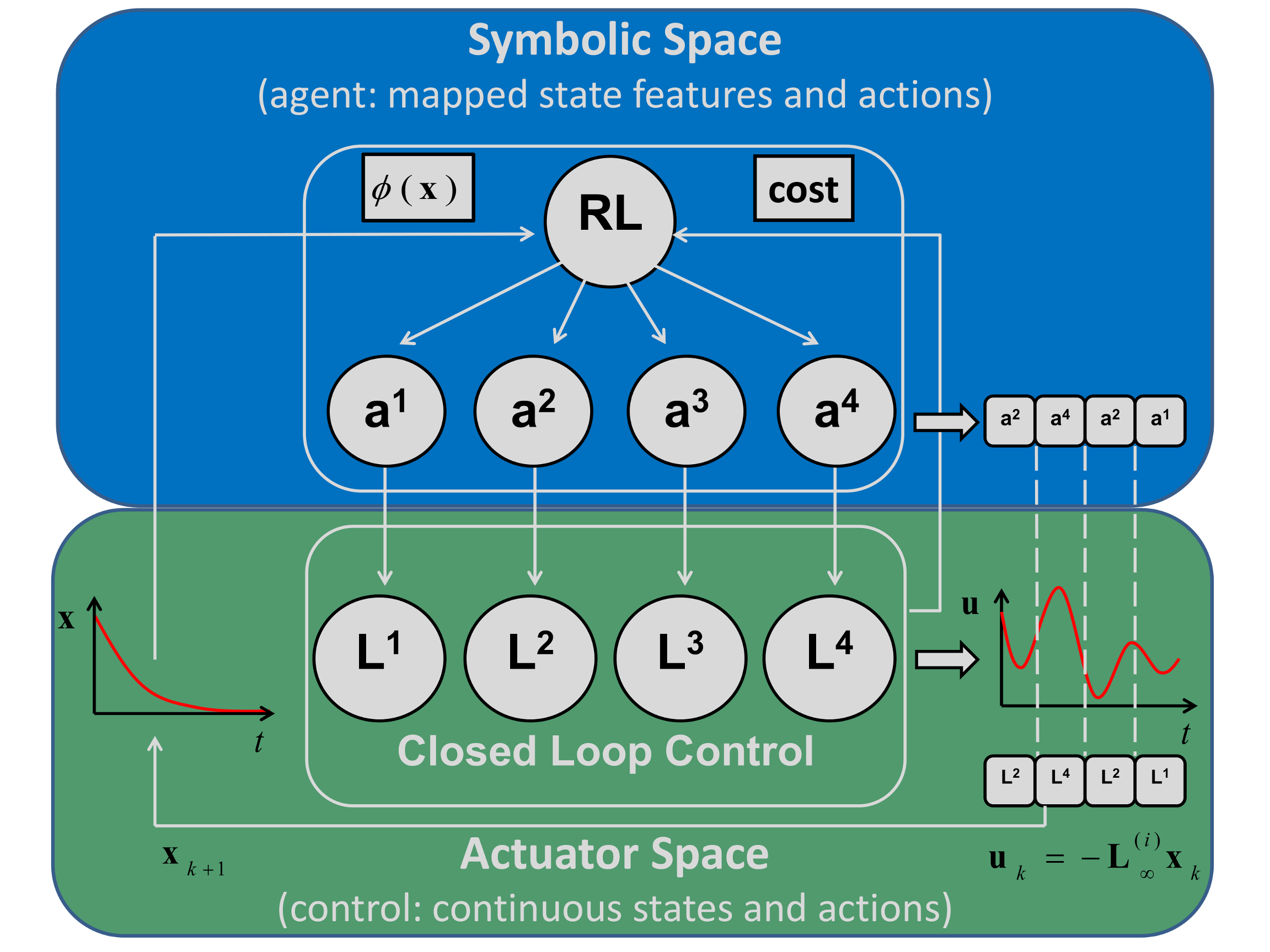}
    	\caption{{\bf Overview of hierarchical RLOC framework.}
    		A top-level model-free RL agent selects between Symbolic Actions $a^{(i)}$, $i=1,\dots,n_{a}$ each linked to an underlying local linear infinite horizon feedback controller $L^{(i)}_{\infty}$, for $i=1,\dots,n_{a}$. 
    		The agent learns a mapping between available controllers and a problem defined state space by chaining actions together, evaluating their outcomes and repeating exploration over many RL episodes. 
    		Each selected feedback gain matrix $L^{(i)}_{\infty}$ is used together with the current continuous state $\bx_{k}$ to produce continuous control $\bu_{k}$ with the action switching dictated by the high-level policy $\pi$.   
    		The closed loop control generates costs: quadratic penalties in state deviation from the target and the size of control used. 
    		The accumulated cost is attributed to the state-action pair $(s,a)$ and this information is passed to the agent, which drives the learning.
    		We define two distinct spaces: \emph{Actuator Space} containing all the low-level (actuator) control information (Optimal Control space, hence it deals with the definition of quadratic cost, continuous states $\bx_{k} \in \reals^{l}$ and control $\bu_{k} \in \reals^{m}$ at time step $k$), and \emph{Symbolic Space} containing all the information relating to the high-level learner (RL agent, assumed to be formulated as an (approximately) Markov Decision Process) which deals with the feature and control mapped variables: discrete states $s^{(i)}, i=1,\dots,n_{s}$ and actions $a^{(i)}, i=1,\dots,n_{a}$, as well as the mapped rewards.}
    	\label{Fig1}
    \end{figure}

    We propose a neurobiologically motivated algorithm, Reinforcement Learning Optimal Control (RLOC), for controlling nonlinear systems with unknown dynamics in an infinite horizon control setting.
    The core idea of the algorithm is a hierarchical structure which uses a high-level symbolic decision maker to control a given nonlinear system by switching between a set of actions, each linked to a low-level linear feedback controller (LQR$_{\infty}$ \cite{Athans1971}). 
    These controllers are learnt on-line through system identification by inference of the linearised plant dynamics at particular points in state space (linearisation points), which are currently randomly positioned (with a view to extend this to unsupervised controller placement). 
    We use system identification to infer an internal forward model of how control signals are coupled to state space, performed with Expectation Maximization for Linear Dynamical Systems with inputs (LDSi) on sufficiently small regions \cite{Ghahramani1996a}. 
    We assume that the local linearisation of the control plant is a "good enough" approximation of the globally nonlinear plant dynamics and propose that RLOC is able to smartly re-use the low-level controllers by optimally engaging them in regions of state space other than where they were obtained. 
    The inferred dynamics are applied in an optimal feedback control formalism to yield LQR$_\infty$ controllers. 
    At the symbolic level the high-level agent learns a global policy of how to best assign each low-level controller to a particular region of the state space so that to solve a  task globally in a single learning run while maximising a long-term reward.
    Thus control in the RLOC framework involves high-level action selection (symbolic selection of the controller) and low-level control (specifying continuous control signals for actuators).
    The high-level policy is improved by employing the low-level quadratic system/control cost, acquired for controlling the plant in each discretised actuator region of state space, as a high-level reward signal, hence closing the learning loop. 
    Therefore, RLOC operates on two levels of abstraction: 
    \emph{Symbolic Space} describing aspects of the symbolic level reinforcement learning agent, such as the discrete actions $\A = \{a^{(i)}:i,\dots,n_{a}\in\naturalnums^{+}\}$ (where cardinality of set $\A$ is finite and equal to the number of linearisation points $n_{a}$), and discrete states $\S = \{s^{(j)}:j,\dots\,n_{s} \in\naturalnums^{+}\}$, where $n_{a}$ and $n_{s}$ are the total number of reinforcement learning actions and states respectively;
    and \emph{Actuator Space} describing aspects of the optimal control problem such as continuous state $\bx \in \reals^{l}$ and continuous control $\bu \in \reals^{m}$, modelled using $k=1,\dots,n_K$ time steps, where $n_K$ is the total number of low-level controller time steps per each RL episode.
    A graphical representation of RLOC's hierarchical architecture is given in Fig \ref{Fig1} and an overview of the learning steps is given in Algorithm \ref{Algo1}.
    
    \begin{algorithm}
    	\fontsize{10pt}{10pt}\selectfont
    	\caption{RLOC Overview \label{Algo1}}
    	\begin{algorithmic}[1]
    		\State Discretise continuous Actuator State Space into Symbolic RL States.
    		\State Collect state-control mapped trajectories using naive control.
    		\For{each controller}   
    		\State Randomly place a linearisation centre in Actuator State Space
    		\State Sample naive state-control trajectories local to the linearisation centre
    		\State Estimate linear model using Linear Dynamical System with inputs
    		\State Derive corresponding linear controller using Optimal Control theory
    		\EndFor
    		\For{a number of RL iterations}
    		\State Use Monte Carlo sampling of the Symbolic States 
    		\State Implement linear controllers as actions  
    		\State Improve RL policy using rewards based on quadratic costs of low-level controllers
    		\EndFor
    		\State Use learnt policy to control the given task from any initial Actuator State.
    	\end{algorithmic}
    \end{algorithm}
    
    More formally, the RLOC algorithm operates by selecting a high-level discrete RL action $a^{(i)}$, which engages a corresponding low-level linear controller $L_{\infty}^{(i)}$, $i = 1, \ldots, n_{a}$. 
    The feedback gain matrix, $L_{\infty}^{(i)}$, represents a single infinite horizon LQR controller, which uses an existing state $\bx_{k}$ to calculate a continuous control $\bu_{k}$ which is in turn applied to the system advancing it to the next state $\bx_{k+1}$ as a response. 
    Hence the continuous time control interacts with the system to evolve the state, generating trajectories of state $\bar{\bx} \in \reals^{l \times n_{K}}$ and control $\bar{\bu} \in \reals^{m \times n_{K}}$.
    These control signals and state features are mapped into RL framework and the learning loop is completed with the inclusion of a low-level quadratic cost $J(\bx,\bu) \in \reals$. 
    This cost is observed by the reinforcement learner and the action selection policy is improved in response.
    
    The low-level feedback controllers $L_{\infty}^{(i)}$ are derived from a collection of approximate linear models of the system $\ldsmodel{i} = \big\{~\zeta^{(i)}, \widehat{A}^{(i)}, \widehat{B}^{(i)}, \widehat{\Sigma}_w^{(i)} ~\big\}$, $i=1,\dots\,n_{a}$, where $\zeta^{(i)}$ is the "linearisation centre" which is a point in state space around which the nonlinear dynamics are linearly approximated, and indicate the neighbourhood in which we would expect the approximation to be most accurate, $\widehat{A}^{(i)} {\in \reals^{l \times l}}$ is the estimated system control matrix, $\widehat{B}^{(i)} {\in \reals^{l \times m}}$ is the estimated control gain matrix and $ \widehat{\Sigma}_w^{(i)}$ is the estimated state covariance noise.
    The initialization of the low-level actions is performed using random placements of the linearisation points in state space. The local linear controllers differentiate from one another based on the nonlinearities of the dynamics around which the linearisation centres are being obtained. We account for the variability resulting from the random controller placements, by performing many learning trials. 
    
    In order to approximate the nonlinear dynamics, i.e. to obtain each $\widehat{A}^{(i)}$ and $\widehat{B}^{(i)}$ matrix, our model interacts with the system using naive control $\bU \in \reals^{m \times  (n_K-1) \times c}$, which is comprised of $c=3^m$ naive control trajectories $\bar{\bu}^{(j)} \in \reals^{m \times (n_K-1)}$, $j=1,\dots,c$, where $m$ is the dimensionality of Actuator Control vector at a single time step $k$.
    Each naive control trajectory $\bar{\bu}^{(j)}$ is applied to each naive starting state $\widetilde{s}^{(i)}, i=1,\dots,n_{\widetilde{s}}$, discretised on a grid, producing state experience $\bX \in \reals^{l \times n_K \times n_N}$, which is comprised of $n_N = c ~n_{\widetilde{s}}$ naive state trajectories collected, $\bar{\bx} \in \reals^{l \times n_K}$, where $l$ is the dimensionality of Actuator State vector at a single time step $k$.
    The state experience $\bX$ is paired with corresponding control trajectories and stored as a 3D-matrix $\bY \in \reals^{(l+m) \times (n_K-1) \times n_N}$.
    
    The collected trajectories $\bY$ are sub-sampled to only give experience local to each linearisation centre $\zeta^{(i)}$, $i=1,\dots, n_{a}$ defined by boundaries on each element of the state's using tolerances $\delta_{\theta}, \delta_{\dot{\theta}}$ (see Appendix Fig \ref{S1_Fig} for an example of such sub-sampling).
    This produces $n_H$ sub-trajectories ${\{(\bar{\bx}_{k:k+h},\bar{\bu}_{k:k+h})_{(i,j)}\}}_{j=1}^{n_{H}}$ for each linearisation centre $\zeta^{(i)}$.
    Each linear model $\ldsmodel{i}$ is then inferred by using parameter estimation for Linear Dynamical Systems with Inputs (LDSi) with the obtained sub-trajectories. 
    The models provide a principled basis of forming the continuous controllers, using the standard infinite horizon LQR theory, hence each high-level action $a^{(i)}$ is linked to a low-level linear feedback controller $L_{\infty}^{(i)}$ corresponding to a learnt linear model $\ldsmodel{i}$, $i=1,\dots, n_{a}$.

    \subsection{Local Linear Approximations} \label{subsec:localLinApprox}
    In order to learn local linear approximations of the dynamics, we perform parameter estimation for the linear dynamical systems with inputs (LDSi) algorithm \cite{Ghahramani1996a}, based on the expectation maximization method described in \cite{ghahramani1996algorithm}.
    Linear Dynamical Systems are used to model the temporal evolution of systems that undergo changes which depend only on the previously observed state. 
    
    An LDSi can be described using the following equations
    \begin{eqnarray}
    \label{LDSi}
    \mathbf{z}_{k+1} &=& A\mathbf{z}_{k} + B\bu_{k} + \mathbf{w}_{k} \label{LDSi1}\\
    \bx_{k} &=& C\mathbf{z}_{k} + \mathbf{v}_{k} \label{LDSi2}
    \end{eqnarray}
    where the observable output, $\bx_{k} \in \reals^{p}$, is linearly dependent on the hidden state $\mathbf{z}_{k} \in \reals^{l}$, while subsequent states, $\mathbf{z}_{k+1}$ are linearly dependent on the prior state $\mathbf{z}_{k}$ and controls $\bu_k$. 
    The terms $\mathbf{w}_{k} \sim \mathcal{N}(\mathbf{0},{\Sigma_{w}})$ and $\mathbf{v}_{k} \sim \mathcal{N}(\mathbf{0},{\Sigma_{v}})$ are the noise for state and observed output respectively, with the diagonal covariance matrices being estimated by the model, $\Sigma_{w} \in \reals^{l \times l}, \Sigma_{v} \in \reals^{p}$. 
    The $A, B$ and $C$ matrices act on the hidden state, input and observable state vectors respectively.
    The LDSi algorithm estimates the $A, B, C$ matrices from data.
    For our simplified model, we assume that the system is non-deterministic, $\mathbf{v}_{k} \not\equiv 0$, with a fully observable state $\mathbf{z}(k) = \bx(k)$, hence $C = I$. 
    The unobserved state noise term accounts for nonlinearities of the system and hence $\mathbf{w}_{k} \not\equiv 0$.
    
     We build a naive control tensor $\bU =(\bar{\bu}{^{(1)}}^\T,\dots,\bar{\bu}{^{(c)}}^\T) \in \reals^{m \times (n_K-1) \times c}$, which contains  individual trajectories $\bar{\bu}^{(j)}\in \reals^{m \times (n_K-1)}$, $j=1,\dots,c$, of harmonically decaying controls, where $m$ is the dimensionality of the control, $n_K$ is the total number of simulation steps of each trajectory and $c$ is the total number of naive control trajectories, using the following steps 
    \begin{itemize}
    	\item Construct $c$ number of vectors $\bu^{(j)}_{1} \in \reals^{m}$, $j=1,\dots,c$, for the first time step $k=1$, where each $m$ dimensional vector is built by choosing $m$ elements from sample points of a naive control set $\D = \{ 0, 1, -1 \}$.
    	We obtain all permutations with replacement by building $c = \left\vert{\D}\right\vert ^m$ such naive control vectors, each containing entries $d_{i,j}$'s, where $d_{i,j}$ is drawn from $\D$. 
    	\item Create $c$ number of matrices $\bar{\bu}^{(j)}\in \reals^{m \times (n_K-1)}, j=1,\dots,c$, by replicating the chosen sample points $d_{i,j}$ along each of the dimensions of each vector for $n_K-1$ steps. 
    	\item Adjust the individual control matrices $\bar{\bu}^{(j)}\in \reals^{m \times (n_K-1)}$, $j=1,\dots,c$, to form harmonically decaying control trajectories, using the following formula for each of the dimensions
    	\begin{equation}\label{eq:naiveControl}
    	{u}^{(j)}_{i,k} = \frac{d_{i,j} ~ b ~  u_{max}} {b + k - 1 }
    	\end{equation}
    	where $k = 1,\dots,(n_K-1)$ time steps, $b$ is the naive control rate of decay, $u_{max}$ is the upper task control bound which remains the same for each of the control dimensions and $d_{i,j}$ is the chosen sample point of $\D$ (note that for a chosen $({i,j})$ pair the sample point remains constant over $k$). 
    \end{itemize}
    
    We perform system identification for a dynamical system around linearisation centres $\zeta^{(i)}$, $i=1,\dots, n_{a}$, using the following steps
    
    \begin{itemize}
    	\item Apply each naive control trajectory, $\bar{\bu}{^{(j)}}, j=1,\dots,c$, (Eq. \ref{eq:naiveControl}) to all naive start states $\widetilde{s}^{(i)}, i=1,\dots,n_{\widetilde{s}}$ (positioned at equally spaced locations on a grid and collect naive state trajectories $\bX = (\bar{\bx}{^{(1)}}^\T,\dots,\bar{\bx}{^{(n_N)}}^\T) \in \reals^{l \times n_K \times n_N}$, where $n_N= c ~n_{\widetilde{s}}$ and $\bar{\bx}^{(j)} \in \reals^{l \times n_K}$.
    	
    	\item Pair the state trajectories $\bar{\bx}^{(j)} \in \reals^{l \times n_K}, j=1,\dots,n_N$ with the corresponding controls $\bar{\bu}{^{(j)}}, j=1,\dots,c$ (the control applied at time step $k$, $\bu{^{(j)}_{k}}$, is paired with the resultant state obtained $\bx^{(j)}_{k+1}$), and store all pairs as "experience" $\bY \in \reals^{(l+m) \times (n_K-1) \times n_N}$.
    	
    	\item Sample $\bY$ for $n_H$ non-overlapping sub-trajectories of length $h$, $\{(\bar{\bx}_{k:k+h},\bar{\bu}_{k:k+h})_{(i,j)}\}_{j=1}^{n_H}$ in the region close to each linearisation centre $\zeta^{(i)}$, for $i=1,\dots, n_{a}$, which we more strictly define by a tolerance on angle and angular velocity deviations from $\zeta^{(i)}$ ($\delta_{\theta}$ and $\delta_{\dot{\theta}}$), where all points of the sub-trajectories are within a bounding box centred at the linearisation centre (see Appendix Fig \ref{S1_Fig}).
    	
    	\item Use the sampled sub-trajectories to estimate the local dynamics $(\widehat{A},\widehat{B})^{(i)}$ for each linearisation centre using parameter estimation for LDSi (see Table \ref{Table1}), where $\bx_{k+1} = I \mathbf{z}_{k+1} + \mathbf{v}_k$ and $\mathbf{z}_{k+1} = \widehat{A}^{(i)} \mathbf{z}_{k} + \widehat{B}^{(i)} \bu_{k} + \mathbf{w}_{k}$ and where the covariance matrices $\Sigma_v$ and $\Sigma_w$ of the $\mathbf{v}_k,\mathbf{w}_k$ noise terms are estimated from data \cite{ghahramani1996algorithm}.
    \end{itemize}
    
    \begin{table}[!ht]
    	\centering
    	\fontsize{10pt}{10pt}\selectfont
    	\caption{\bf{Parameters for linear model inference.}}
    	{\begin{tabular}{|l|l|l|l|}
    			\hline
    			\bfseries Parameter Description & \bfseries Symbol & \bfseries Value & \bfseries SI Unit\\ \hline  
    			LDSi number of cycles & $n_c$ & 100 & [$a.u.$]\\ \hline
    			LDSi likelihood tol. & $\delta_{tol}$ & 0.0000001 & [$a.u.$]\\ \hline
    			Sub-trajectory tol. angle deviation from $\zeta^{(i)}$ & $\delta_{\theta}$ & 20 & [$deg$]\\ \hline
    			Sub-trajectory tol. angular vel. deviation from $\zeta^{(i)}$ & $\delta_{\dot{\theta}}$ & 120 & [$deg/s$]\\ \hline
    			Naive control rate of decay & $b$ & 10 & $[a.u.]$ \\ \hline
    	\end{tabular}}
    	\begin{flushleft}
    		Full set of parameters necessary for inference of linear models $\ldsmodel{i}$ for both benchmark tasks.
    	\end{flushleft}
    	\label{Table1}
    \end{table}
    
    For a given model $\ldsmodel{i}$, it is reasonable to assume that the linear model built using $(\widehat{A},\widehat{B})^{(i)}$ is an accurate approximation close to $\zeta^{(i)}$, and also that an LQR controller $L^{(i)}_{\infty}$ gives a good control strategy close to this point, for $i=1,\dots, n_{a}$.
    Hence each local linear model $\ldsmodel{i}$ allows a principled way of deriving feedback gain matrices $L^{(i)}_{\infty}$ corresponding to high-level actions $\{a^{(1)},a^{(2)},\dots\,a^{(n_{a})}\}$. 
    We predict that RLOC can learn how to reuse the low-level controllers in areas of state space other than where they were obtained. 
    This mapping is able to be learnt by RLOC due to its hierarchical structure and the ability of the high-level decision maker to operate on an abstract domain.

    \subsection{Low-Level Controller Design} \label{subsec:lowLevelCntrDesign} 
    In engineering terms we want to solve a broad class of nonlinear control problems, with a continuous time evolution of the system dynamics 
    \begin{equation} \label{eq:xdot}
    \dot{\bx} = \mathbf{f}(\bx,\bu), ~\bx_{0}=\bx^{init}
    \end{equation} 
    where $\bx^{init}$ is the initial state.
    The type of discrete time nonlinear systems we choose to learn (which are an approximation of the continuous time equivalent), have the following system dynamics
    \begin{equation}\label{eq:sysdyn}
    \bx_{k+1} = A(\bx_k) \bx_k + B(\bx_k) \bu_k, ~\bx_{0}=\bx^{init}
    \end{equation}
    where the state $\bx_{k} \in \reals^{l}$ and control $\bu_{k} \in \reals^{m}$ at time step $k = 1, \ldots, n_{K}$ (where $n_{K}$ is the total number of simulation steps) act via $A(\bx(k)) \in \reals^{l \times l}$, the state dependent system control matrix, and $B(\bx(k)) \in \reals^{l \times m}$, the state dependent control gain matrix. 
    The aim is to find a sequence of controls $\bar{\bu}$ that minimises the squared state and control error costs over infinite time horizon.
    If $A(\bx(k))=A$ and $B(\bx(k))=B$ are constant over the whole state space, then the problem reduces to the well-known Linear Quadratic Regulator (LQR) problem $\bx_{k+1} = A\bx_k + B\bu_k $. 
    
    In general, tasks require a non-zero target state, ${\bx}^{\ast}$, meaning that the objective is to move the system to, and keep it at, that target state, i.e. attain $\bx_k = \bx_{k+1} = \bx^{\ast}$. For notational convenience, and reflecting the inherent assumption of a zero target in the Linear Quadratic Regulator (LQR) theory \cite{doya2007bayesian}, we define a change of variables $\bt x(k) = \bx(k) - \bx^{\ast}$, such that in this shifted coordinate system, the target is at zero, i.e. $\bt x^{\ast}=\mathbf{0}$. 
    The linear dynamics are then defined in this shifted coordinate system in the form
    \begin{equation} \label{eq:linDynOffset}
    \bt x_{k+1} = A\bt x_{k} + B\bu_{k}
    \end{equation} 
    We can now define the cost function as
    \begin{equation} \label{eq:CostDefinition}
    J(\bt x,\bu) = \frac{1}{2} \sum_{k = 1}^{\infty} \bt x_k^\top W \bt x_k + \bu_k^\top Z \bu_k
    \end{equation}
    with symmetric positive semi-definite state weight matrix $W\in \reals^{l \times l} \geq 0$ and symmetric positive definite input (control) weight matrix $Z\in \reals^{m \times m} > 0$, specifying the penalty on the state deviation from target and the penalty on the control size respectively. 
    The final time step $n_K$ weight matrices are imposed by $W_{n_K}$ and $Z_{n_K}$. 
    In the infinite horizon setting used here, these are equal to the matrices applied at incremental time steps $W=W_{n_K}$ and $Z=Z_{n_K}$. 
    LQR seeks to minimise the cost function $J(\bt x,\bu)$ which captures the objective of moving the system to the target in a manner defined by the weight matrices.
    
    We wish to solve the class of problems with an infinite time horizon and assume the optimal value function to be quadratic.
    The Bellman equation is minimised with respect to control, producing an analytical result for the infinite horizon optimal control law 
    \begin{equation}\label{eq:optu}
    \bu_k = - L_{\infty} \bt x_k
    \end{equation}
    where $L_{\infty}\in \reals^{m \times l}$ is the infinite horizon optimal feedback control gain matrix, which is approximated with a finite horizon control via iterating backwards in time the quadratic cost-to-go function (the value function) ${V} {\in \reals^{ l \times l}}$ for a large number of iterations $\widetilde{n}_{{K}}\gg 10^3$, $k = \widetilde{n}_{{K}}, \widetilde{n}_{K-1},\ldots, 1$, and $V_{\widetilde{n}_{{K}}} = W$ \cite{kwakernaak1972linear}
    \begin{equation}\label{eq:Vfunction}
    {V}_k = {W} + {A}^\top V_{k+1}{A} - {A}^\top V_{k+1}{B}({Z}+{B}^\top  V _{k+1}{B})^{-1}{B}^\top V_{k+1}{A}
    \end{equation}   
    For a large $\widetilde{n}_{K}$, the feedback gain matrix $L_1$ converges to the infinite horizon gain matrix $L_\infty$, $\lim_{\widetilde{n}_{{K}} \to \infty} L_{1} \to L_\infty$.
    It is called the static compensator - since it does not vary with time, and therefore can be calculated off-line using the following equation, obtained after minimizing the Bellman equation w.r.t. control
    \begin{equation}\label{eq:Lmatrix}
    L_{\infty} \approx {L_{1}} = ({Z}+{B}^\T{V_{2}}{B})^{-1}{B}^\T{V_{2}}{A}
    \end{equation} 
    
    Note that the system dynamics are not made available to the agent a priori and are instead approximated around a few (currently randomly) selected points using experience via local linear approximations with LDSi, while the penalty imposed on the inputs make the controller more robust to model inaccuracies \cite{Andreson1971}.
    So, for each linearisation point $\zeta^{(i)}$, RLOC learns a local model of the dynamics with ($\widehat{A}^{(i)}, \widehat{B}^{(i)}$), and uses this to calculate, off-line, the feedback gain matrices $L^{(i)}_{\infty}$ (referred to as the $i$-th controller). 
    Each controller, $L^{(i)}_{\infty}$, in turn, corresponds to a high-level reinforcement learning action $a^{(i)}$.
    
    The cost matrices are specified for each task individually and establish a penalty on state and control in order to design appropriate feedback gain matrices.  The size of the penalties provide a guide on the importance of each state's dimension is to the overall goal. 
    A general penalty on the size of the plausible force was imposed for all of the tasks in order limit large forces that could be exhibited by the controller. 
    For each tasks a hard cut-off control size ceiling was imposed in an expert manner: a biologically plausible robotic arm control does not exceed 10 [Nm] and the cart-pole forces were not allowed to exceed 20 [N]. 
    For the two benchmark tasks presented here, the cost matrices were selected as: 
    (a) robotic arm state and control matrices respectively $W$ = diag(30, 30, 0, 0), $Z$ = diag(1,1), i.e. end effector's positions were penalised for deviating from the target joint configuration, while no penalties were imposed on angular velocities; 
    (b) cart-pole state and control matrices respectively $W$ = diag(30, 3, 2000, 200), $Z$ = (1), i.e. the cart was penalised for deviating from its pre-defined stopping position on the track and the pole was penalised for deviating from its upright target position. 
    The cart-pole system requires to be actively balanced even when the target is reached, due to the target being an unstable equilibrium, however the overall need to drive the system to a stationary upright position is imposed by moderate penalties on velocities. 
    
    \subsection{High-Level Controller Design} \label{subsec:highLevelCntr}
    In reinforcement learning, an agent repeatedly interacts with its environment, which is typically formulated as a Markov Decision Process, MDP, (a discrete time stochastic control process, which is defined by its state and action sets, and by one-step dynamics of the environment)
    by choosing actions and observing the resulting state and reward, reflecting desirability of the most recent state change.
    The agent's objective is to maximize the aggregated reward signal over time.
    Model-free reinforcement learning typically makes no assumption about the system dynamics other than the Markov property, which requires subsequent state and reward to depend only on previous state and action, and optimises control purely through interaction with the environment.
    The RLOC algorithm is hierarchically structured, which means that the original low-level Actuator Space is Markovian, since the environment's response at $t+1$ depends only on state and action at $t$, more formally $\Pr\{ s_{t+1}=s', r_{t+1}=r | s_t, a_t \} ~\forall s', r, s_t, a_t$. 
    However the use of abstractions in creating the high-level controller, results in some loss of information, thus making the Symbolic Space semi-Markovian \cite{singh1994learning}. 
    Even though the Markov property is not strictly met, the Symbolic Space is a coarse approximation to that, and our method of choice for the high-level controller, Monte-Carlo RL, is known to be robust in situations where this property is not strictly met \cite{sutton1998reinforcement}. 
    
    Standard model-free RL algorithms for discrete state problems include Monte-Carlo (MC) Control, SARSA and $\Q$-learning \cite{sutton1998reinforcement}.
    We choose to employ $\varepsilon$-greedy on-policy MC Control, in which the agent interacts with the environment by rolling out trajectories, recording the $\{r,s,a\}$ triplets and maximising a long term reward.
    We use a model-free algorithm because it is a light-weight method compared to model-based RL, both in terms of the development/computation time and space requirements.
    In this paper we aim to show that RLOC works as a proof-of-principle and model-free MC algorithm provides a feasible solution for this.
    Monte-Carlo is also a natural algorithm choice, not only because it deals well in a semi-Markovian environment, but also because the tasks addressed in this paper break down into sequences of repeated trials (of pole balancing / arm reaching), having a defined goal and a notion of a final time step, making them episodic, which is a requirement for the MC algorithm. 
    Arguably MC is better suited to tasks under episodic conditions than SARSA or $\Q$-learning, because MC does not bootstrap, i.e. do not make estimates based on other estimates, and keeps all the information about the visited states without the need for approximation.
    We choose on-policy algorithm, since we can interact directly with the tested systems and update the value function based on experience.
    The use of $\varepsilon$-greedy policy with respect to $\Q^\pi$ allows maintaining a degree of exploration of the non-greedy actions during learning and is guaranteed to be an improvement over any other $\varepsilon$-soft policy, $\pi$, by the policy improvement theorem \cite{sutton1998reinforcement}. 
    It has been known to outperform the $softmax$ action selection  \cite{Kuleshov2014} and is considered as the preferred method of choice \cite{Heidrich2009}.
    We choose iterative (vs. batch) MC in order to take advantage of "recency", the updating the policy with the most recent experience after every epoch.
    This also alleviates computational requirements since information is used straight away without storing it over many batches.
    
    The $\varepsilon$-greedy on-policy MC algorithm keeps a table of action-values $\widehat{\Q}^{\pi}(s,a) \in \reals^{n_{s} \times n_{a}}$, under policy $\pi$, where $n_{s}$ is the number of Symbolic (RL) States and $n_{a}$ is the number of Symbolic (RL) Actions, with entries for each state-action pair $(s,a)$, and aims to approximate the expected future discounted return
    ${\Q}^{\pi}(s,a) = {\bf E}_{\pi}(\R_{t} | s_{t}=s,a_{t}=a) = {\bf E}_{\pi}(\sum_{n=0}^{\infty} \gamma^{n}r_{t+n+1} | s_{t}=s,a_{t}=a)$, where $t$ is the RL time step, $\gamma\in [0,1]$ is the discount factor and $r$ is the observed immediate reward.
    The control decisions are made by the policy $\pi$, which maps past experience to actions, and can be probabilistic.
    For simplicity, we focus on stationary policies, where this mapping depends only on the current state.
    So for an observed state $s$ and action $a$, $\pi(s,a)\in [0,1]$ gives the probability of choosing $a$ in $s$.
    The critic only reinforcement learning uses the estimated $\Q$-values to determine the policy and in particular we are interested in the $\varepsilon$-greedy policy, where given estimates $\widehat{\Q}^{\pi}(s,a)$ $\forall s,a$, the policy $\pi(s_{t},a_{t})=\Pr(a_{t}|s_{t})=1-\varepsilon$ iff $a_{t}\!=\!\argmax_{a}\widehat{\Q}^{\pi}_{t}(s_{t},a)$ and $\frac{\varepsilon}{|\A|-1}$ otherwise, $\varepsilon \in [0,1]$. 
    
    In our implementation of the algorithm, a step-size parameter $\alpha_i = \frac{\widetilde{\alpha}}{i^{\mu}}$ (i.e. $\widetilde{\alpha}$ is reduced based on the number of $i$ visits to the $(s,a)$, where $\mu$ is the rate of decay, is used to incrementally improve the estimate of the action-value function, $\widehat{\Q}^{\pi}(s,a)$, during epoch $\tau^{(j)}$, $j=1,\dots,n_\Tr$.
    The $i$-th update of state $s$, using action $a$ (i.e. $i$-th visit to state action pair $(s,a)$)  corresponds to the $t$-th time step of the $j$-th epoch
    
    \begin{eqnarray} \label{eq:Q_update}
    \widehat{\Q}_{i+1}(s,a) &=& \widehat{\Q}_{i}(s,a) + \alpha_{i} \big(\R_{i}(s,a) - \widehat{\Q}_{i}(s,a)\big)\\
    \R_{i}(s,a) &=& \sum_{n=0}^{n^{(j)}_{T} - t} \gamma^{n}r_{t+n+1}
    \end{eqnarray}
    where $\R_{i}(s,a)$ is the $i$-th $(s,a)$ visit return while in current epoch $\tau^{(j)}$, obtained by averaging discounted immediate rewards from time step $t$ of epoch $j$ to the end of epoch time step ${n^{(j)}_{T}}$, corresponding to the $i$-th update of $(s,a)$.
    Each MC epochs $\tau^{(j)}$ is simulated for $n_K$ number of low-level controller time steps and whenever a new symbolic state $s^{(i)}, i=1,\dots,n_{s}$ is entered the agent either chooses a new or the same action $a^{(i)}, i=1,\dots,n_{a}$ (which engages the corresponding $L_{\infty}^{(i)}, i=1,\dots,n_{a}$ LQR controller). 
    We perform every-visit Monte Carlo, meaning that all of the visits to $(s,a)$ pairs are considered for a given epoch (see Table \ref{Table2} for full list of the discussed parameters).
    \begin{table}[!ht]
    	\centering
    	\fontsize{10pt}{10pt}\selectfont
    	\caption{\bf{RLOC Learning Parameters}}
    	{\begin{tabular}{|l|l|l|l|}
    			\hline
    			\bfseries Parameter Description & \bfseries Symbol & \bfseries Value & \bfseries SI Unit\\ \hline  
    			MC epsilon & $\epsilon$ & 0.1 & [$a.u.$]\\ \hline
    			MC epsilon decay & $\nu$ & 0.1 & [$a.u.$]\\ \hline
    			MC gamma & $\gamma$ & 1 & [$a.u.$]\\ \hline
    			MC alpha & $\tilde{\alpha}$ & 1 & [$a.u.$]\\ \hline
    			MC alpha decay & $\mu$ & 0.5 & [$a.u.$]\\ \hline
    			MC number of epochs & $n_\Tr$ & 2000 & [$a.u.$]\\ \hline
    			Number of experiment trials & $n_{\widetilde{N}}$ & 500 & [$a.u.$]\\ \hline
    			Number of Symbolic Actions & $n_{a}$ & $\{ 1,\dots,10 \}$ & [$a.u.$]\\ \hline
    	\end{tabular}}
    	\begin{flushleft}
    		Full set of reinforcement learning parameters and LDSi parameters used for both benchmark tasks.
    	\end{flushleft}
    	\label{Table2}
    \end{table}
    
    The high-level reward $r$ obtained for using action $a^{(i)}$ in state $s^{(i)}$ at time step $t$, is the negative cumulative low-level control cost for entering the Actuator State $\bx_{k}$ (corresponding to symbolic state $s^{(i)}$) at time step $k$ and applying continuous control until the system first leaves that symbolic state at time step $k'$, at which stage the epoch triplet $\{r_{t+1},s_t, a_t\}$ is recorded for RL time step $t$. 
    The total number of RL steps $n_{T}$ varies between epochs, depending on the number of state-action pairs visited. 
    
    In summary, the RLOC algorithm begins each epoch roll-out by a random choice of Symbolic State $s^{(i)}, i = 1,\dots,n_s$ (corresponding to an initial Actuator State $\bx^{(i)}_0$) and a random choice of action $a^{(i)}, i = 1,\dots,n_a$ (corresponding to a feedback gain matrix $L_{\infty}^{(i)}$) to be used in that Symbolic State. 
    At each actuator time-step $k$, the current high-level action $a^{(i)}$ engages the corresponding low-level continuous controller $L_{\infty}^{(i)}$, which uses the current Actuator State $\bx_k$, to determine the approximately optimal continuous Actuator Control $\bu_k$ (see Eq. \ref{eq:optu}), and the system progresses to the next state $\bx_{k+1}$ as a consequence.
    The epoch simulation lasts for $k = 1, \dots, n_K$ low-level control time steps, during which a number of controller switches occurs, based on the number of Symbolic States visited, giving $t$ reinforcement learning time steps.
    
    We define a \emph{control mapping} where $\psi(a^{(i)}) \rightarrow L^{(i)}_{\infty}$ for $i=1,\dots,n_{a}$
    \begin{equation} \label{controlMapping}
    \psi:\A\to\reals^{l\times m}
    \end{equation}
    and a \emph{feature mapping} is defined between a Actuator State $\bx_{k}$ and a Symbolic State $s \in \S$, where $\phi(\bx_{k}) \rightarrow s^{(i)}$ for $i=1,\dots,n_{s}$
    \begin{equation} \label{featureMapping}
    \phi:\reals^{l}\to \S
    \end{equation} 
    The reinforcement learning agent is used to learn the best action choice at each Symbolic State to minimise the cost-to-go, giving a policy $\pi:\S\to\A$.
    The control and feature mappings can be combined with a learnt deterministic policy, $\pi$, to define the full \emph{control policy} $\widehat{\omega}:\reals^{l}\times\naturalnums\to\reals^{m}$, which is an approximate estimate of the time-dependent optimal control mapping $\omega^{\ast}$
    \begin{equation}
    \label{PolicyDefinition}
    \omega^{\ast}(\bx_k, k) \approx \widehat{\omega}(\bx_{k},k) = \psi(\bx_{k},\pi(\phi(\bx_{k})))\bx_{k}
    \end{equation}
    
    This completes the formal definition of each level of our hierarchical framework RLOC, and the overall algorithm is presented in Algorithm \ref{Algorithm2}.
    We now present results obtained by applying RLOC to two benchmark nonlinear control problems.
    
    \begin{algorithm} 
    	\fontsize{10pt}{10pt}\selectfont
    	\caption{RLOC Algorithm \label{Algorithm2}}
    	{
    		\begin{enumerate}\setlength{\itemsep}{0pt}\setlength{\topsep}{0pt}
    			\item Discretise $n_l \leq l$ chosen dimensions of the Actuator State, $\bx \in \X$, into Symbolic States $\{s^{(i)},s^{(2)},\dots\,s^{(n_{s})}\}$ to obtain equally spaced $n_l$-dimensional grid.
    			\item Collect $(\bX, \bU)$ trajectories using naive control sequences ("motor-babbling").
    			\begin{algorithmic}[1]
    				\For{each controller $a^{(i)},i=1,\dots,n_a$,}
    				\State Randomly place linearization centre $\zeta^{(i)}$ in state space $\X$, $i=1,\dots\,n_{a}$
    				\State Sample $n_H$ sub-trajectories $\{(\bar{\bx}_{k:k+h},\bar{\bu}_{k:k+h})_{(i,j)}\}_{j=1}^{n_H}$ local to $\zeta^{(i)}$
    				\State Estimate linear model $\ldsmodel{i}$ with LDSi
    				\State Derive infinite horizon feedback gain matrix $L^{(i)}_{\infty}$ using LQR theory
    				\State Associate symbolic action $a^{(i)}$ with feedback gain matrix $L^{(i)}_{\infty}$
    				\EndFor
    			\end{algorithmic} 
    			\item Initialize policy and state-action values, for all $s\in \S$, $a \in \A(s)$. 
    			\begin{algorithmic}[1]
    				\State $\pi(s,a)$ $\leftarrow$ $\frac{1}{\left\vert{\A}\right\vert}$ 
    				\State $\Q^{\pi}(s,a)$ $\leftarrow$ $\mathbf{0}$
    			\end{algorithmic} 
    			\item Approximate best policy with Monte-Carlo RL.
    			\begin{algorithmic}[1]
    				\For{epoch ${\tau^{(i)}, ~ i = 1,...,n_\Tr}$} 
    				\State Pick a random state $s \in \S$, choose its centre as initial Actuator State $\bx_{0}$
    				\State Sample action $a_0 \sim \Pr(a|\pi, s_0)$ and set symbolic time $t=0$
    				\State Initialise control $\bu_0$ from initial Actuator State $\bx_0$ and controller $\psi(a_0)$
    				\For {for Actuator time step $k=1,\dots,n_K$}
    				\State Update $\bx_k$ using previous state $\bx_{k-1}$, control $\bu_{k-1}$ and system dynamics.
    				\If {state changes, e.g. $\phi(\bx_k) \neq s_t$}
    				\State Store total quadratic cost for action $a_t$ in state $s_t$ as reward $r_{t+1}$
    				\State Append $(r_{t+1},s_t, a_t)$ triplet to existing trace
    				\State Increment symbolic time, $t = t+1$
    				\State Update state, $s_t = \phi(\bx_k)$
    				\State Sample action $a_t \sim \Pr(a|\pi, s_t)$
    				\EndIf
    				\State Update control $\bu_k$ from current state $\bx_k$ and controller $\psi(a_t)$
    				\EndFor
    				\State Update $\widehat{\Q}^{\pi}(s,a)$ $\leftarrow$ $\widehat{\Q}^{\pi}(s,a) + \alpha_{i} \big(\R_{i}(s,a) - \widehat{\Q}^{\pi}(s,a)\big)$ $\forall ~s$ in epoch
    				\State Update $\pi$ from $\widehat{\Q}^{\pi}$
    				\EndFor  
    			\end{algorithmic}
    			\item Finalise policy $\pi(s)$ $\leftarrow$ $\argmax_{a} \widehat{\Q}^{\pi}(s,a)$ $\forall s \in S$.
    			\item Use $\pi$ to control the system from any Actuator State to the target. 
    		\end{enumerate}
    	}
    \end{algorithm}

    \section{Results}
    \subsection{Benchmark Systems}
\begin{figure}[!ht]
    	\centering
    	\includegraphics[width=0.75\textwidth]{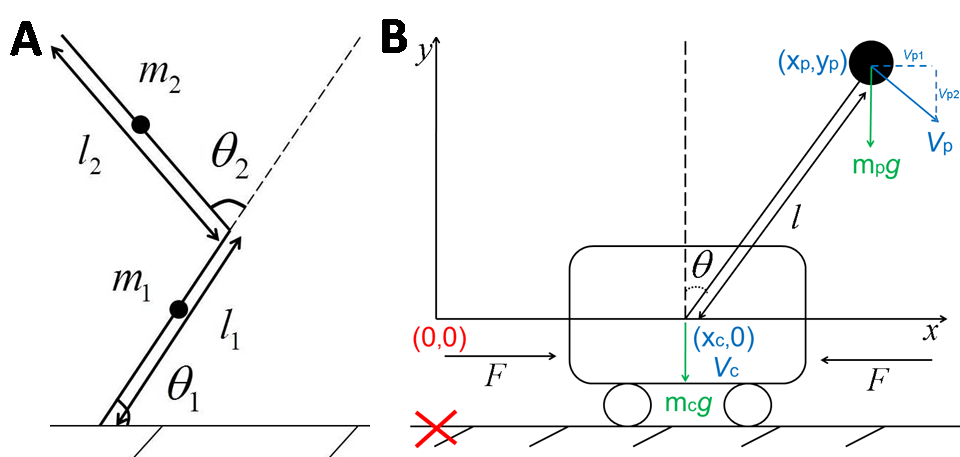}
    	\caption{{\bf Nonlinear control benchmark problems.}
    		(\textbf{A}) Planar 2D robotic arm is modelled as 2 weightless rods of lengths $l_1$, $l_2$ and masses $m_1$, $m_2$ centred near the middle of each link. 
    		The shoulder joint is 'attached' to a point, while allowing both the shoulder and elbow joints to move freely at angels $\mathbf{\theta} = (\theta_{1},\theta_{2})^\T$ positively defined to the right, and restricted to a range $(0,180\degree) \in \reals^2$.
    		A biologically plausible torque constraint of $[-10,10]$ Nm is applied to each joint.
    		(\textbf{B}) A freely swinging pendulum is  attached to a cart on a track: the task is to stabilize the pendulum in an upright and stationary position with the cart stopping at a predefined place on the track (marked with a red cross and located at the cart position $z=0$).
    		The pendulum angle $\theta$ is defined positively to the right, $l$ denotes the length of the pendulum, ${m}_p$ and ${m}_c$ are the masses of the pendulum and the cart, respectively, and $F$ denotes the force applied to the front/rare of the cart, which is limited to $[-20,20]$ N. 
    		The system has a stable equilibrium for $\theta\in\{-180\degree,180\degree\}$ (pendulum hanging down), and the target state is the unstable equilibrium at $(\theta,\dot{\theta}) = \mathbf{0}$ (pendulum stationary and upright).
    	}
    	\label{Fig2}
    \end{figure}

    We demonstrate our proof-of-principle algorithm's effectiveness on two standard nonlinear control problems: the 2-link robotic arm (RA, see Fig \ref{Fig2} \textbf{A}) and the cart-pole (CP, see Fig \ref{Fig2} \textbf{B}) systems, whose dynamics are not available to the algorithm a priori.
    The RLOC algorithm learns locally linear dynamics at randomly positioned linearisation points from naive control experience on-line, while the simulated environment is being perturbed by additive white Gaussian system noise. 
    Both of the benchmark tasks are standard in the field of control theory, with the cart-pole system typically used as a standard low-dimensional benchmark task for testing performance of control algorithms.
    Abbeel et.al. \cite{Duan2016} have aided in the process of setting up well-defined benchmarks for nonlinear continuous control, and suggest that the cart-pole should be a standard low-dimensional benchmark problem for comparing deep RL algorithms.
    The planar robot arm reaching movements are used as an approximation to studying human motor control in reaching tasks, and the task is extensively studied in control algorithms \cite{todorov2005generalized,li2004iterative}, robotics and are relevant to neuroprosthetics.
    For example, Lonsberry  et.al. \cite{lonsberry2017deep} study 2 DOF robot arm actuated with 6 muscles in order to use computerized control for providing an alternative to neural regeneration via electric muscle stimulation.
    Both of the benchmark tasks are therefore relevant to testing the performance of our control algorithm, RLOC, and we select them to demonstrate at first instance our algorithm's effectiveness and low computational requirements, as a proof-of-principle framework. 
    
    The robotic arm system consists of 2 links operating in a plane, with the shoulder joint 'attached' to a point, while allowing both the shoulder and the elbow links to move freely in the constrained state space.
    The robotic arm is represented using the state $\bx = (\theta_1,\theta_2,\dot{\theta}_1,\dot{\theta}_2)^\T \in \reals^{4}$ and the first time derivative $\dot{\bx} = (\dot{\theta}_1,\dot{\theta}_2,\ddot{\theta}_1,\ddot{\theta}_2)^\T \in \reals^{4}$, with the \emph{forward dynamics} 
    $
    \boldsymbol{\ddot{\theta}}
    = 
    {M}(\boldsymbol{\theta})^{-1}
    (\boldsymbol{\tau} - {C}(\boldsymbol{\theta},\boldsymbol{\dot{\theta}}) - {B}\boldsymbol{\dot{\theta}})
    $
    where $\boldsymbol{\theta}\in \reals^2$ is the joint angle vector, $\boldsymbol{\dot{\theta}}\in \reals^2$ is the joint angular velocities vector, ${M}(\boldsymbol{\theta}) \in \reals^{2 \times 2}$ is a positive semi-definite symmetric inertia matrix, ${C}(\boldsymbol{\theta}, \boldsymbol{\dot{\theta}})\in\mathbb{R}^2$ is a vector of Centripetal and Coriolis forces, ${B}\in\mathbb{R}^{2\times2}$ is the elbow joint friction matrix and $\boldsymbol{\tau} = \bu = (u_1,u_2)^\T \in \reals^2$ is the joint torque defined to be the system control. 
    The robotic arm control task represents planar arm reaching movements, starting from any initial state configuration $\boldsymbol{\theta}^{init} \in [0,180\degree]$, $\dot{\boldsymbol{\theta}}^{init} \in [{-250\degree},250\degree]$ performed optimally to a selected target $\bx^{\ast}$ by applying controls (torques) $\bu \in [-10,10]$ Nm to each joint using policy $\boldsymbol{\pi}(s)$ (target chosen as $\bx^\ast = [90\degree,90\degree,0,0]$).
    
    The second system we study is the swing-up-and-stop version of the cart-pole problem: the task consists of stabilising a free swinging pendulum attached to a cart moving on a track, with the cart stopping in a predefined position on the track and the pendulum upright and stationary (we call this the "target" state). 
    This is considered a standard, difficult benchmark task in the field of nonlinear control theory since the system is underactuated and inherently unstable, hence any learning algorithm e.g. \cite{doya2000reinforcement,todorov2005generalized,Peters2008}, is required to deal with both the swing-up and the stabilisation phases where pole needs to be actively balanced to remain upright. 
    Note, only one LQR controller (obtained by linearising the system at its target state) can only solve this problem effectively from states in the vicinity of the unstable equilibrium of the upright position.
    The cart-pole system is represented with $\bx = (z,\dot{z},\theta,\dot{\theta})^\T \in \reals^{4}$ and $\dot{\bx} = (\dot{z},\ddot{z},\dot{\theta},\ddot{\theta})^\T \in \reals^{4}$ vectors, with the \emph{inverse kinematics}
    $
    {M}(\mathbf{q}){\ddot{\mathbf{q}}}^2 + {C}(\mathbf{q},{\dot{\mathbf{q}}}){\dot{\mathbf{q}}}+ {G}(\mathbf{q}) + {B}\widetilde{\dot{\mathbf{q}}} = \boldsymbol{\tau}
    $
    where $\boldsymbol{\tau} = (F,0)^\T \in \reals^{2}$, where $F \in \reals$ is the force applied to the system, ${\mathbf{q}} = [z,\theta]^\T \in \reals^{2}$, $\dot{\mathbf{q}}=(\dot{z},\dot{\theta}) \in \reals^{2}$, $\widetilde{\dot{\mathbf{q}}} = (\mbox{sgn}(\dot{z}),\dot{\theta})^\T \in \reals^{2}$, where $\mbox{sgn}(\cdot)$ means 'sign of', $\ddot{\mathbf{q}}=(\ddot{z},\ddot{\theta}) \in \reals^{2}$, ${M}(\mathbf{q}) \in \reals^{2 \times 2}$ is a positive definite symmetric matrix, ${C}(\mathbf{q},{\dot{\mathbf{q}}}) \in \reals^{2 \times 2}$, ${G}(\mathbf{q}) \in \reals^{2}$ and $B=(B_c,B_p)^\T$ is a vector of friction parameters.
    The cart-pole control task represents applying the correct amount of force $\bu \in [-20,20]$ N to the cart either from the front or rear side of the cart along the horizontal plane, in order to reach the unstable upright equilibrium and stop the cart in a pre-defined position on the track (i.e. target of $\bx^{\ast} = \mathbf{0}$).
    For more information on these two systems, parameter setting and simulation techniques see Appendix Sections \ref{S1_Appendix} and \ref{S2_Appendix}.

    \subsection{RLOC Performance}
\begin{figure}
    	\centering
    	\includegraphics[width=0.75\textwidth]{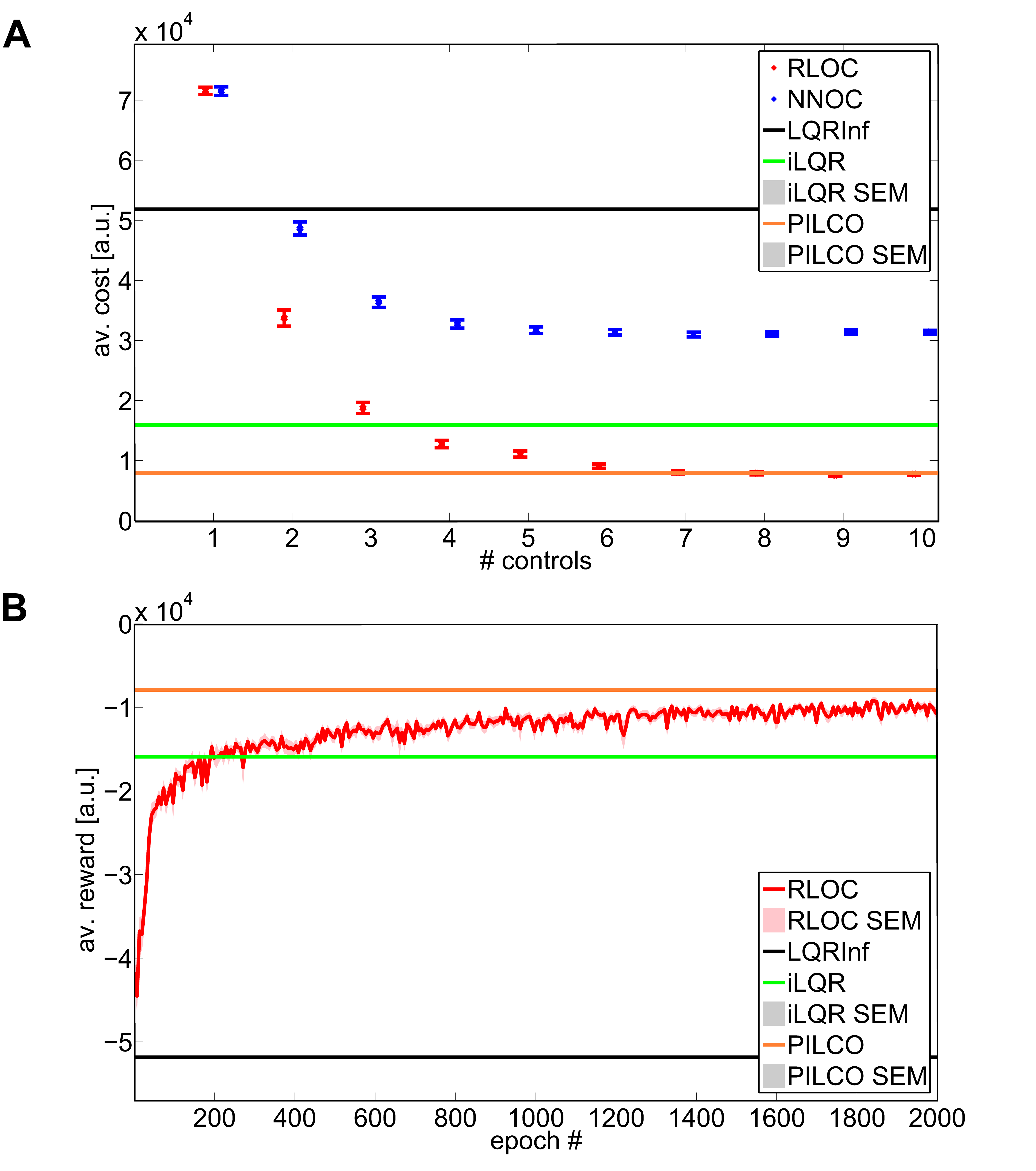}
    	\caption{
    		{\bf Learning curves (cart-pole).} 
    		({\textbf{A}}) The performance of RLOC control solution learnt as a function of individual controllers (data points) with respect to engineering state of the art solutions (solid lines). 
    		The av. cost for controlling the cart-pole from equally spaced 100 start positions to unstable equilibrium end-point, is computed for RLOC and the benchmark PILCO (orange line displayed as mean and standard error of the mean of 3 learning trials for each starting position), iLQR solver (green line displayed as mean and standard error of the mean of 3 learning trials for each starting position) and the infinite horizon LQR solver (black line displayed as the mean average cost of the deterministic control of 1 learning trial per each starting position, with each linearisation point located at the start position).   
    		The NNOC (blue data points displayed as mean and standard error of the mean of 500 learning trials) shows performance for the cart-pole system without high-level learning, which involves switching to the nearest linearisation point (controller). 
    		The RLOC (red data points with learning trial-to-trial variability displayed as mean and standard error of the mean of 500 learning trials) shows performance for a system with both high-level and low-level learning. 
    		({\textbf{B}}) We choose a number of controllers for which RLOC's average cost is stable (here chosen to be 8) and investigate the shape of the learning curve for the 500 trials. 
    		The plot shows the average "direct reward sample" (defined as the current full epoch's cumulative reward experienced by the algorithm at a fixed 6 epoch interval during 2000 epoch learning trial), taken across corresponding epochs of the 500 trials and reported as mean and standard error of the mean, colour coded in red and pastel red respectively.
    		The benchmark algorithms PILCO (orange), iLQR (green) and LQR infinity (black) are shown as solid lines.
    	}
    	\label{Fig3}
    \end{figure}

    The performance of RLOC for controlling the cart-pole system is shown in Fig \ref{Fig3} and compared to standard benchmark average costs of infinite horizon LQR (black solid line), iLQR (green solid line) and PILCO (orange solid line) averaged through repeated trials as tested on the same machine.
    The control algorithms LQR$_{\infty}$ and iLQR were simulated for both of the benchmark tasks for $n_K$ total number of time steps, while PILCO was trained using 100 time steps (total simulation time of 5 seconds with discretisation step  $\delta t$ = 0.05). 
    The quadratic cost function used to train the LQR$_\infty$, iLQR and RLOC were identical, whereas PILCO was trained using the saturating cost (typically employed by the algorithm) but evaluated on the quadratic cost function for consistency.
    The iLQR algorithm was initialised with best guess control forces $\bu_{0}$ sampled from a set $\{-1, 0, 1\}$ as advised by the authors \cite{li2004iterative}.
    The benchmark costs were obtained by averaging the individual start state costs of attempting to reach the upright unstable equilibrium target from 100 equally spaced start positions in Actuator State Space: the LQR infinity used as many number of controllers as starting states (i.e. 100 linearisation points, located at the centre of each start state) and used 1 trial since it is deterministic, the iLQR used as many controllers as the number of simulation time steps (with each control trajectory recalculated from scratch for individual start-target states, calculated recursively to achieve convergence for 3 trials), and PILCO used 50 Gaussian basis functions to give a probabilistic policy for each state (obtained on state-by-state basis and repeated for 3 trials).
    Note: PILCO average cost for one learning trial was found by averaging the cost for controlling the cart-pole from each of the 100 starting states (using final learnt policy obtained after 5 iterations of PILCO algorithm). 
    This mean cost was then averaged across the 3 repeated trials and reported as mean and standard error of the mean.

    The average RLOC cost for controlling the cart-pole from 100 equally spaced start positions to a predefined upright, stationary target as a function of the number of controllers available to the system is depicted in Fig \ref{Fig3} {\textbf{A}}. 
    To account for  the variability in learning performance resulting from the initial random placement of the controllers, we repeated the random placement and subsequent learning (we refer to this as one learning "trial") 500 times.  
    The average RLOC cost for one learning trial was obtained by taking a mean of the control cost (after learning) across all start configurations of the cart-and-pendulum (100 possibilities). 
    Following this the per-trial cost was averaged across 500 random initialisations of the controllers placements and reported as the average cost and standard error of the mean across trials. 
    To make the results comparable across the increasing number of controllers obtained via random placement of linearisation points, we grew the number of controllers used in the reinforcement learning agent, $n_a = \{1,\dots, 10 \}$, by preserving the previously obtained linearisation locations at $(n_a-1)$ and adding additional controller sequentially until maximum number of tested linearisation points was reached (repeated for 500 trials).  
    The RLOC data points show a smooth decrease in the average cost as the number of controllers increases.
    We show that on average RLOC is able to outperform deterministic state-of-the-art control system iLQR, and match the mean solution cost of probabilistic state-of-the-art algorithm PILCO, while achieving a stable performance with using only 7 controllers for the cart-pole task.
    We demonstrate that the high-level learning for smart controller switching is necessary by plotting the average cost associated with naive nearest neighbour controller switching using policies obtained by RLOC, versus an increasing number of controllers during the 500 trials.
    The Naive Nearest Optimal Control (NNOC) involves switching to the nearest controller during task execution by calculating the Euclidean distance from the current Symbolic State to all of the available linearisation points and picking the one with the shortest Euclidean distance, therefore providing non-smart selection between locally linear controllers of the low-level learner.
    The results show that NNOC average cost is higher than that of RLOC which proves the necessity for the high-level learning. 
    It is worth remarking that the average cost of RLOC for 1 controller is different from the LQR infinity mean cost because RLOC uses one controller located at the target state, while the LQR infinity controllers are located in the centre of each starting state (i.e. using 100 controls in total). 
    Even with the random controller placement RLOC is able to find optimal solutions for the cart-pole problem with fewer controllers and with less cost than state-of-the-art control solutions, specifically it beats iLQR 96$\%$ of the time and it beats PILCO 75$\%$ of the time in terms of cost (see Appendix Fig \ref{S2_Fig}).
    To visualise results we operated RLOC with 8 controllers (which provides a stable performance) and plot an example learning curve. 
    
    The learning curve achieved by the RLOC algorithm is shown in Fig \ref{Fig3} \textbf{B}, and plotted as a direct sample of the learning reward signal (experienced by the reinforcement learner during the 2000 epochs learning and sampled at 6 epoch interval) averaged through 500 trials and reported as mean and standard error of the mean. 
    The plot shows that on average, the algorithm starts off with a performance better than the LQR infinity (black solid line) and is able to outperform state-of-the-art iLQR algorithm (green solid line) after 250 epochs of learning. 
    The reinforcement learning policy used to control the cart-pole during learning is probabilistic hence retaining a degree of exploration of less desirable actions, therefore the reported direct reward mean is just below that of PILCO, compared to the deterministic policy used to obtain (cf. A) where the RLOC policy finds the same solution as that of PILCO.
    
\begin{figure}
    	\centering
    	\includegraphics[width=0.75\textwidth]{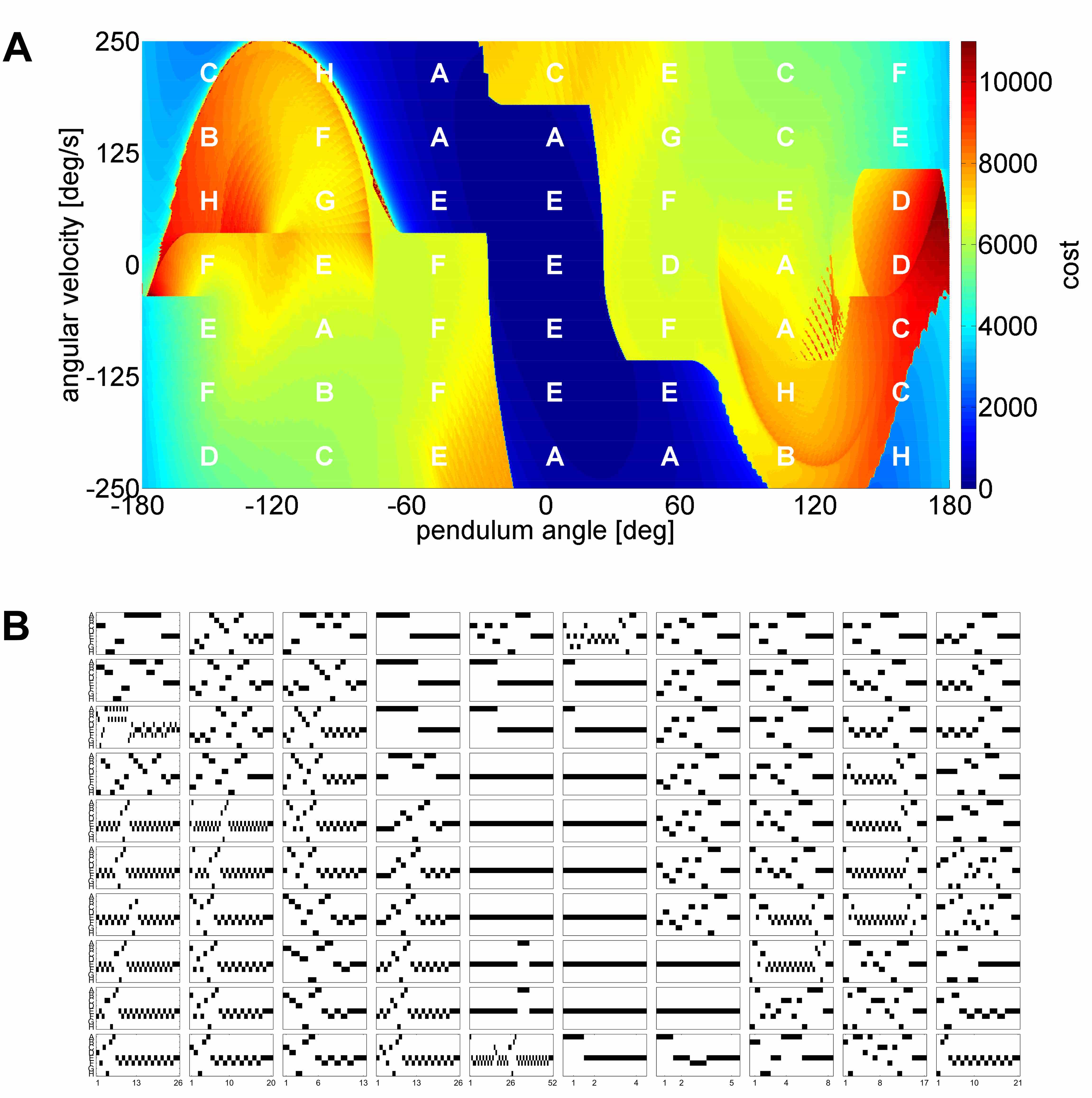}
    	\caption{
    		{\bf Policy evaluation (cart-pole).} 
    		({\textbf{A}}) Example RLOC value function (cost-to-go) plot depicting control to state space mapping, exhibiting the algorithm's learnt choice of actions when a particular area of space is encountered during task execution. 
    		For the purposes of demonstration a random final policy using 8 actions was selected from the 500 learnt policies of the executed trials.
    		The plot demonstrates the cost of the learnt deterministic control for starting in any area of state space (discretised to $10^6$ states, learning for 2000 epochs, using 8 actions labelled A - H, any of which can be assigned by the high-level policy to any of the 49 available Symbolic States) and controlling the task to unstable upright equilibrium end-point, with warmer areas indicating a higher incurred cost. 
    		The final action-to-state space assignment learnt by the high-level policy is shown with letters superimposed onto an area of state space where the action is active, while the shape of the displayed cost function is dictated by the underlying dynamics of the task. 
    		({\textbf{B}}) The action sequences obtained during control of the cart pole to upright target from 100 start states while using the same policy used to obtain (A).
    		Each row within the sub-plot indicates an action (labelled A-H) which, if chosen, is marked with a black dot. 
    		Each column corresponds to the cart-pole moving to a new Symbolic State, where a new action (which may be the same) is chosen.
    		The initial controller of each start state corresponds to the action assigned by the policy for that area of state space as displayed in (A). 
    		The figure shows that RLOC's high-level policy dictates the linking of actions into sequences and demonstrates to not only be able to reuse low-level actions, but also to learn reusing whole concatenated action sequences for solving the task.
    		Mostly, topologically conjunct regions of state space produce similar patterns of activation, which are clearly correlated with regions of the value function, while the switching in the action patterns can be seen to occur as the value function abruptly changes.
    	}
    	\label{Fig4}
    \end{figure}

    The evaluation of a learnt policy for the cart-pole task using 8 controllers (labelled A - H) from 100 equally spaced starting states is shown in Fig \ref{Fig4} .
    Every time the system enters a new Symbolic State, the high-level policy dictates the action, corresponding to an underlying LQR infinity controller, which should be engaged in that state, thus generating sequences of actions as the cart-pole is guided to its upright target. 
    Fig \ref{Fig4} \textbf{A}  displays the state space plot of the final value function, expressed using costs for performing the control task from $10^6$ start positions with 2000 RL epochs and a 10 second simulation of the cart-pole system.
    The linearisation centre for each controller of the example policy has the general form $\zeta^{(i)} = [0, 0, \theta^{(i)}, 0]^\T$, and in the specific example policy used in the above figure, the pendulum angle had the following randomly selected entries $\theta^{(i)} = \{0, -155\degree, -109\degree, -86\degree, -10\degree, 86\degree, -160\degree,  155\degree\}$, for controllers $i = \{ 1,\dots,8 \}$  respectively.
    The colormap indicates the difficulty of the task with respect to its dynamics, while the superimposed letters refer to the low-level continuous control policy mapping from actions to states. 
    The cart-pole value function plot corresponds well to the optimal value function obtained using Dynamic Programming method that has full knowledge of the dynamics (see Fig 2a, Deisenroth et al., \cite{Deisenroth2009a}) and exhibits the expected discontinuity near the central diagonal band.
    The band is given by those states from which a small cost is incurred for bringing the pendulum up to the unstable equilibrium target without dropping it first.  
    The cart-pole is an underactuated system, which means that the pole cannot be directly controlled to achieve arbitrary trajectories, thus for the majority of the state space first enough momentum needs to be built up in the swung up stage before stabilisation can begin (while both of these phases accumulate cost). 
    Therefore, beginning from the upright unstable equilibrium, once the starting angle position goes beyond critical point, an abrupt increase in the cost is observed, which is due to the pole first being dropped (regardless of the force applied to the cart) and then being swung back up.
    The value function largely displays symmetry reflecting the periodic nature of the task, however the learnt high-level RLOC action-state mapping (and hence the displayed cost-to-go) is not necessarily symmetrical due to the nature of the abstract learning. 
    
    Distinct sequences of controller actions ("action sequences") obtained during control of the cart-pole system are shown in Fig \ref{Fig4} {\textbf{B}}.
    Each row depicts the activation time of one controller (black dot, labelled A - H). 
    Each column corresponds to the cart-pole moving to a new Symbolic State where a new action (which may be the same) is chosen.
    The amount of switching depends on the cart-pole trajectory through state space, since the policy is allowed to engage a new controller once a new Symbolic State is entered.   
    The continuous motor control  commands generated by RLOC exhibit characteristic action patterns across state space. 
    Regions of state space can be seen to produce similar, characteristic patterns of action sequences during task execution. 
    Mostly topologically conjunct regions of state space produce similar patterns but on the boundary between some regions abrupt changes in patterns occur. 
    These regions, while characterised by the sequence of actions, are clearly correlated with regions of the value function and switching in action patterns occurs as the value function abruptly changes, see Appendix Fig \ref{S3_Fig} for analogous robotic arm results.
    
\begin{figure}
    	\centering
    	\includegraphics[width=0.75\textwidth]{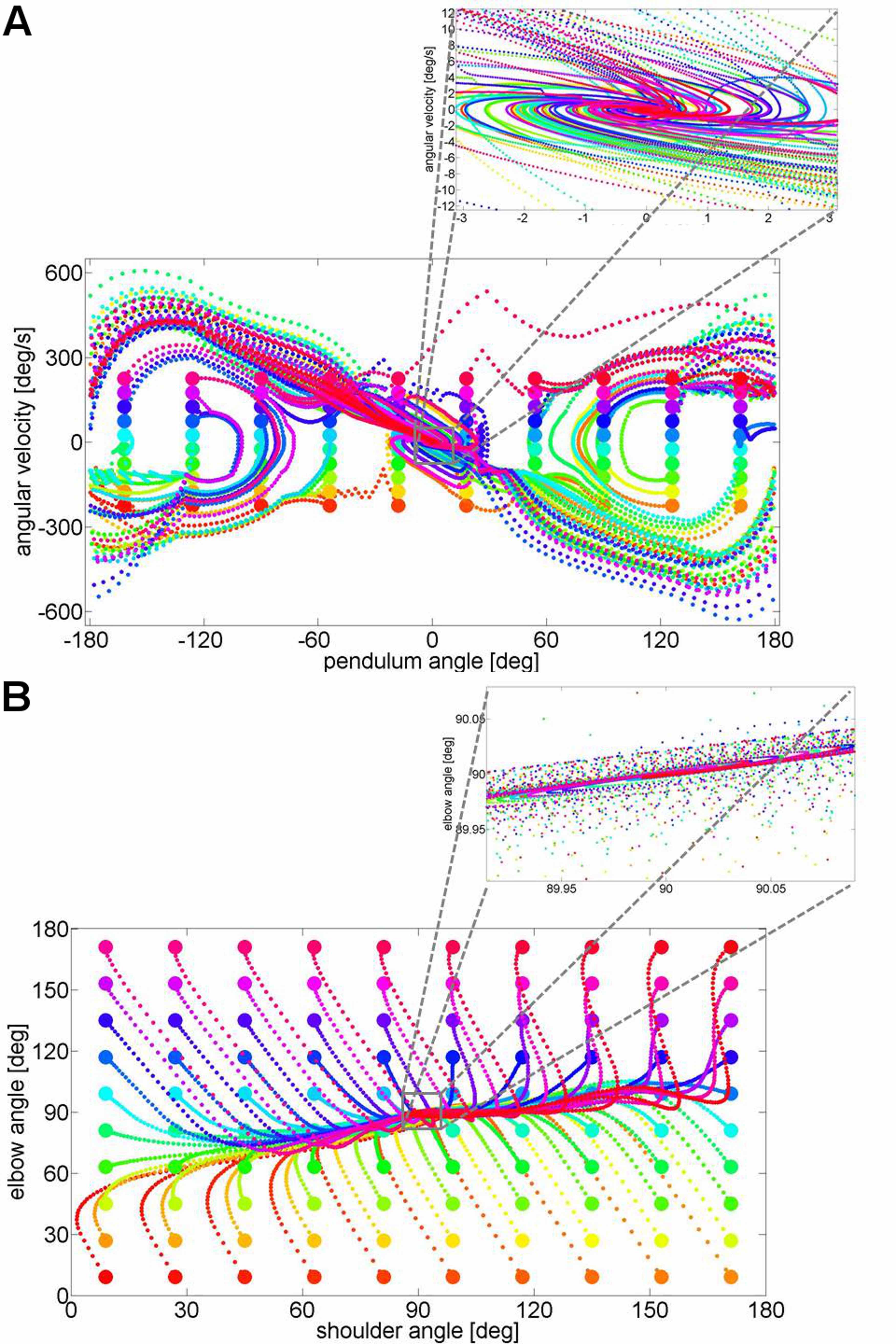}
    	\caption{
    		{\bf Example trajectories for learnt policies (cart-pole, robotic arm).} 
    		({\textbf{A}}) The figure depicts cart-pole trajectories obtained using the learnt policy. 
    		The zoomed in plot shows the behaviour of the pendulum near the target. 
    		({\textbf{B}}) The example state trajectories for the robotic arm obtained using control of a learnt RLOC policy.
    	}
    	\label{Fig5}
    \end{figure}

    The characteristic trajectories experienced during control of the cart-pole and the robotic arm systems are depicted in Fig \ref{Fig5}.
    These were evaluated with the same policies used to obtain the final value functions of each task for a period of 10 sec (i.e. approximating an 'infinite horizon' control setting, since this period is longer than the learning phase, 3 sec) was used to investigate the behaviour of the system beyond the learnt simulation time.
    This allowed us to observe how the system copes with prolonged task execution, showing that the RLOC policy is able to continue to control the system after learning phase is complete, over an infinite horizon and how the algorithm stabilises the system following unexpected perturbations.
    This is in contrast to state-of-the-art iLQR technique which calculates the required control for a specific start-end state combination and for a pre-specified number of time steps only. 
    The displayed trajectories demonstrate that the RLOC learnt policy is successfully able to control the tasks to the predetermined target using smooth control (showed in the zoomed panels) from any area of the phase space (tested in a grid with 100 start states denoted by solid circles of different colour).
    
    The trajectories obtained from controlling the periodic cart-pole task, represented on a flat state space surface with a discontinuity shown at the stable equilibrium of the downward pole position, which means that visually single trajectories are sliced at that point, even though they are continuous (-180$\degree$ is equivalent to +180$\degree$), see Fig \ref{Fig5} \textbf{A}. 
    The figure demonstrates that most start states require the pendulum to build up enough momentum first, before being swung up towards unstable upright equilibrium which is consistent with the nature of the underactuated cart-pole task. 
    The RLOC algorithm successfully stabilises the pendulum in an upright and stationary position with the cart stopping in a predefined position on the track for all 100 starting states of the Symbolic  Space, as shown in the zoomed in section containing the location of the target.
    Even though the target state requires both the pole and the cart to be stationary, meaning that the cart and the pendulum velocity is driven to zero by the high-level control, some degree of movement is required to be retained in order to actively balance the pole in its unstable equilibrium (as seen by the spiral trajectories near the target area). 
    
    The shoulder and elbow joint trajectories for the robotic arm, resulting from the applied control under RLOC final policy, are shown in Fig \ref{Fig5} \textbf{B}.  
    All of the 100 starting states are successfully controlled in a smooth manner to the target located in the centre of the depicted state space. 
    In contrast to the cart-pole system, the robotic arm was controlled to perform planer reaching movements, therefore the arm is not required to work against gravity, hence RLOC can be seen (cf. zoomed in box section) to control both the shoulder and elbow to an accurate near stop position for all of the 100 starting states, as expected.
    The 10 seconds simulation time for both experiments is longer than the 3 second simulation time allowed for the learning phase, demonstrating that the learnt policy can be used for an 'indefinite' amount of time after the learning phase is complete.
    The systems were tested for the response of RLOC control to unexpected perturbations and we observed that both systems were returned to target (visual plots not reported here). 
    The ability of RLOC to control a chosen system from any area of state space is highly advantageous, since for most control algorithms (e.g. iLQR) the control sequences must be recalculated for different start-end points, while RLOC defines a single policy after just one complete learning trial covering the entire state space. 
    
    \section{Discussion}
    Many real world control problems have unknown dynamics, which makes it difficult to apply standard control engineering solutions. 
    Moreover, typical control problems require efficient solutions, i.e. those that achieve a goal while minimising a cost, thus requiring a form of optimality in their control which is difficult to compute for nonlinear task domains. 
    Hence, optimal or near-optimal solutions to nonlinear control problems with unknown dynamics remain an open challenge in both control, robotics, and machine learning fields \cite{todorov2005generalized,kappen2011b}.  
    
    \textbf{Neurobiological Considerations}
    Consider that biological systems learn to manipulate their actuators and external objects with ease, and that this ability has been optimised by competitive drive of evolution over millions of years.
    Thus we looked to biological systems for inspiration and proposed a neurobiologically motivated hierarchical framework for optimally learning non-linear control tasks with unknown dynamics.
    Resent research shows that on the level of actuators high-level motor commands are mapped to movement as evidenced by correlated muscle activation \cite{d2003combinations} and limb movement patterns \cite{Thomik2012}. 
    Linear feedback controllers have emerged in the theories of optimal motor control \cite{scott2004optimal}, with evidence suggesting that our brain learns optimal control for object manipulation through system identification of the dynamics of the arm and object. 
    Correspondingly RLOC implements low-level linear optimal feedback controllers that are learned through local system identification. 
    At the level of optimal action selection in symbolic serial decision making tasks, the results show that humans, monkeys and mice produce behaviour and brain signals consistent with known model-free and model-based reinforcement learning algorithms \cite{d2008bold,glascher2010states}.
    Thus we proposed to hierarchically combine  a high-level reinforcement learning symbolic action representation with low-level continuous controllers.
    The crucial part that enables RLOC to learn nonlinear control tasks using both high-level and low-level control in an integrated manner is the closed loop feedback error propagation, which is comprised of the low-level controllers' cost function, quadratic in state and control, which drives adaptation and improvement of the endogenous high-level reinforcement signals.
    RLOC high-level controllers can be thought of as abstract controllers that can generate specific series of high-level actions towards the completion of a complex task goal (as highlighted in Fig \ref{Fig1}). 
    We show how this results in stereotypically structured action sequences which abstract away from specific initial conditions of the task but group control actions into generic patterns, following overall rules but local deviations, as depicted in the robotic arm control task (see Appendix Fig \ref{S3_Fig} {\textbf{B}}) and the cart-pole control task (see Fig \ref{Fig4} {\textbf{B}}). 
    This is an important finding which could aid in transfer learning, where a policy learnt on one task is applied to a similar but different task, which is therefore not learnt from scratch \cite{stolle2006policies}.
    
    A number of neurobiologically inspired control algorithms exist in the literature and we explain how RLOC's hierarchical framework differs conceptually from other hierarchical or modular approaches to control:
    (a) Todorov et al. \cite{Todorov2005a} proposed a hierarchical optimal control framework for control of nonlinear redundant systems, which was inspired by the hierarchical divide-and-conquer strategies that seem to be implemented in the nervous system. 
    The general idea is that in sensorimotor control tasks, the brain monitors a small number of task parameters and generates abstract commands (high-level control) which are mapped to muscle activations using motor synergies (low-level control). 
    We note that the function performing high-D state to low-D state mapping was expertly provided by the authors and was specific to investigation of the 6 muscle planar arm model, while the system dynamics were known a priori.
    This contrasts RLOC, which does not use any expert systems and does not have a priori knowledge of the dynamics. 
    (b) Another seminal framework, Modular Selection and Identification for Control (MOSAIC), developed by Haruno et al., addresses an accurate generation of motor control by creating a modular architecture where all modules simultaneously contribute to learn a combined weighted feedforward motor command. 
    The weights are derived from the predictions of an internal forward model, actual and desired state. 
    After this modular weighted combination however, the motor command is combined with an unspecified feedback controller to learn object tracking tasks \cite{haruno2001mosaic}. 
    The MOSAIC framework uses supervised learning in order to understand how each module contributes to the overall motor command, with the algorithm having a flat (non-hierarchical) structure across the $n$-paired units producing pure motor control commands. 
    We note that RLOC operates at a different level of representation and utilises a hierarchical approach: it uses a reinforcement learning agent as a high-level decision maker for action selection and action sequencing, which allows abstraction and dimensionality reduction as compared to pure low-level motor control. 
    Some of the neuroscience research used as inspiration for the RLOC algorithm should be taken with caution: the validity of Optimal Feedback Control theory as basis of motor control \cite{Nagengast2009} has been questioned with criticisms of it being a mathematical construct those cost function can be tuned in order to fit behaviour and thus cannot be disproven \cite{Loeb2012}.
    It is argued that the value used for deriving optimality is an attribute of states that are caused by movement which is a consequence not a cause \cite{Friston2011}.
    
    \textbf{Benchmark Performance}
    We demonstrated the system's effectiveness and efficiency of its performance on two standard low-dimensional benchmark non-linear optimal control problems: 2D robot arm control and swing-up and balancing of the pole on a cart while stopping in a pre-defined position on the track.  
    Specifically RLOC matches performance or outperforms algorithms in a number of following ways.
    The hierarchical nature of RLOC design allows the algorithm to benefit from using high-level learning, which provides abstraction and task complexity reduction, by efficiently structuring control of high-level actions to achieve a complex task goal.
    Each high-level action engages a low-level locally linear optimal controller (linearised at a specific point in state space). 
    In the naive solution (tested baseline) the system switches automatically to the nearest linear optimal controller as it moves through state space. 
    We show that high-level action selection has over 50$\%$ lower average cost in finding optimal solutions than the naive solution, in solving the cart-pole task (see Fig \ref{Fig3} \textbf{A}).
    The combination of a high-level action reinforcement learner with continuous low-level optimal controllers allows a significant reduction in the number of controllers typically necessary as compared to the non-learning iterative optimal control algorithm iLQR, by approx. a factor of $10^2$ linearised controllers.
    Fig \ref{Fig3} \textbf{A} shows that on average 4 controllers are sufficient for RLOC to solve the non-linear optimal control problem of the cart-pole upswing with the same final cost as the 300 controller equivalents (recalculated iteratively) as used by iLQR. 
    On the other hand we also reduce the overall state complexity as compared to stand-alone reinforcement learning (operating without approximations) methods by a factor of approx. $10^3$: the cart-pole task RLOC requires approx. $2\times10^2$ state-action pairs to be explored vs. $2\times10^5$ as required by the flat RL algorithm. 
    Moreover we show that on average our algorithm (no prior knowledge of dynamics, random controller placements) outperforms the state-of-the-art probabilistic control system PILCO (no prior knowledge of the dynamics) 75$\%$ of the time, and state-of-the-art deterministic control algorithm iLQR (full prior knowledge of the dynamics) 96$\%$ of the time, when using 7 or more controllers (see Appendix Fig \ref{S2_Fig}), when solving the cart-pole task.

    \textbf{Data Efficiency}
    RLOC is a data efficient algorithm: it learns both the low-level dynamics as well as the high-level control so that within 250 trials (taking $1.25$ minutes running time) 
    it matches the performance of iLQR (see Fig \ref{Fig3} \textbf{B}), and requires an order of $4\times10^2$ less trials when compared to learning a similar pendulum task performed in \cite{yoshimoto2005acrobot}. 
    In contrast, an algorithm using a monolithic continuous RL agent \cite{doya2000reinforcement}, was reported to have taken 1000 epochs (approx. $6$ hours computation time) to learn the benchmark cart-pole task, controlled from any area of state space, without reporting the learning curves. 
    We compare RLOC to PILCO, the current state-of-the-art and the most data efficient learning control algorithm which uses complex representations based on Gaussian Processes to leverage uncertainty of the learnt data.
    This algorithm is trained through forward modelling \cite{rasmussen2008probabilistic,deisenroth2011pilco} (requiring only $17.5$ seconds of interaction with a physical system to learn the benchmark cart-pole problem), however it requires intensive off-line calculations for global policy parameter optimisation -- it takes a reported order of $1$ hour off-line computation time for a single start-target state learning on the cart-pole task \cite{rasmussen2008probabilistic}). 
    The recent advancement of PILCO work is based on a moment matching approach using probabilistic model predictive control \cite{Kamthe2017}.
    It improves on PILCO's need to require the full planning horizon and the need for off-line optimisation of a large number of parameters per control dimension. 
    The algorithm takes $9$ seconds of physical system interaction time (as compared to PILCO's $17.5$ seconds) when solving the cart-pole balancing task, and is reported to improve on the off-line calculation time without providing the order of improvement.
    In comparison, RLOC has a simple representation enabling fast computation to learn nonlinear control from any start position in state space to a target position in a single learning run, while performing all the necessary calculations on-line (and on current computers) much faster than real-time. 
    Comparing the computational requirements of RLOC and PILCO for solving the cart-pole task for 100 different start states of the pole, we find that RLOC takes approximately $2$ minutes, while PILCO algorithm takes approximately $10$ hours as timed on the same machine. 
    This is in part because PILCO uses probabilistic representations of uncertainty that are expensive to compute, specifically Gaussian Processes require an inversion of the kernel matrices which scales $\mathcal{O}(n^3)$ unfavourably with respect to the number of data points required for learning e.g. when the task is temporally-complex or has a very high dimensional state space, although approximative solutions may improve the performance.  
    Meanwhile, RLOC benefits from using linear optimal controllers for which the feedback gain matrices are not state dependent and hence are calculated off-line once, with the most expensive operation of $\mathcal{O}(n^3)$, where $n$ is the dimensionality of the state vector. 
    The high-level learning is performed iteratively using Monte Carlo RL algorithm which is not as expensive (scaling as $\mathcal{O}(n^2)$). 
    
    \textbf{RLOC Architecture}
    RLOC learning operates at two levels of abstraction: RL for the optimal sequencing of actions and LQRs for continuous optimal control of action execution, to achieve a composite task goal.
    The two levels are integrated naturally by RLOC as the low-level control cost (quadratic in state and control) determines the reward signal generated as a result of choosing the high-level actions. 
    This is in contrast to Hierarchical RL algorithms, where the system switches to a different sub-task (action) once its sub-goal has been reached. 
    These sub-goals have to be either specified and defined by hand or may be learnt on-line. 
    In contrast, RLOC action selection chooses an underlying local optimal controller and switches to another one from the top-down without a predefined temporal and/or sub-goal structure.
    We suggest that our architecture may be beneficial for small embedded computing agents aiming to apply nonlinear control, due to its much lower computational cost. 
    
    The RLOC algorithm is a proof-of-principle framework, which general function does not depend on the specific algorithms used to select action sequences or to perform low-level continuous control.
    We have previously shown \cite{Abramova2012} that a more generic set of control policy primitives can be used by RLOC and tested a mixture of LQR and Proportional Integral Derivative (PID) controllers as low-level controllers.
    It is possible to provide a wide pool of controllers to the high-level agent since it can learn to select only those controllers which are best suited to each area of state space, thus discarding any sub-optimal controllers as necessary.
    We have found that PID controllers proved useful in controlling areas of state space, thus supporting the above claim \cite{Abramova2012}. 
    Even though the hierarchical RLOC structure substantially reduces the curse of dimensionality problem, it does not eradicate it, and eventually the problem of exponential state-action pair exploration growth as the number of states increases would limit RLOC's ability to learn highly dimensional systems.
    This is to be addressed through future work, by using function-approximation methods for reinforcement learning thus moving to continuous high-level state and action representations, that could also blend between controllers \cite{DaSilva2009} instead of discretely switching between them.

    \section{Conclusion}
    We proposed a hierarchical reinforcement learning algorithm which learns the control of nonlinear systems with unknown dynamics and demonstrated its efficiency and effectiveness by comparing learning results to two state-of-the-art control algorithms: iterative linear quadratic regulator and probabilistic inference for learning and control, using standard low-dimensional nonlinear benchmark problems.
    RLOC was able to rival/beat the performance of these nonlinear state-of-the-art learning policies \cite{li2004iterative,rasmussen2008probabilistic}, as tested using a single cost function on the same machine for the same benchmark problems. 
    
    RLOC substantially advances previous work on algorithms with a similar hierarchical structure using a high-level reinforcement learner which engages low-level continuous LQR controllers \cite{randlov2000combining,yoshimoto2005acrobot}, and it does so in a number of ways:
    (a) as opposed to relying on the knowledge of system dynamics, RLOC learns from on-line experience and infers only the information necessary in order to form $n$ linear controllers and is thus able to deal with systems with unknown dynamics;
    (b) RLOC learns the low-level LQR controllers on-line, which is in contrast to expertly pre-defined LQR controllers which where specifically tuned to the cart-pole task and had specific switching conditions established in advance;
    (c) as opposed to single start-end control, RLOC learns a global control policy for controlling a given system from any area of state space to a given target in a single learning run; 
    (d) RLOC reduces the number of necessary training passes in order to learn a task, as compared to the previous work \cite{randlov2000combining,yoshimoto2005acrobot}. 
    
    RLOC demonstrates that the basic idea of assembling a global control policy from more primitive fragments through reinforcement learning is a viable framework for solving nonlinear control problems with unknown dynamics.

    \bibliography{paper.bib}
    \clearpage
    
    \appendix
    \appendixpage
    \addappheadtotoc
   	\section{Example Sub-sampling of the Collected Naive Trajectories}
\begin{figure}[h!]
   		\centering
   		\includegraphics[width=0.75\textwidth]{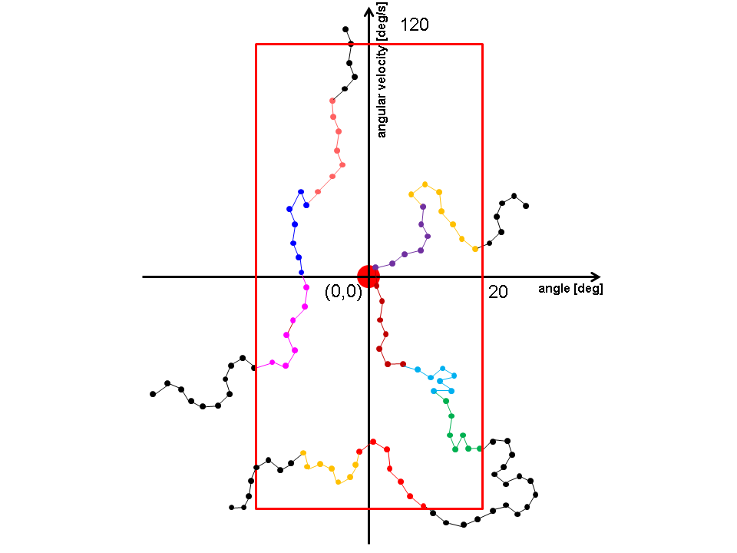}
   		\caption{
   			{\bf Example sub-sampling of the collected naive trajectories.} 
   			Example depicting sampling of the cart-pole collected naive trajectories $\bY$ in order to obtain $n_H$ sub-trajectories, using 2 dimensions of the state trajectory $\bar{\bx}$, the pendulum angle $\theta$ (x-axis) and pendulum angular velocity $\dot{\theta}$ (y-axis).
   		}
   		\label{S1_Fig}
   	\end{figure}

   	Figure \ref{S1_Fig} shows how ten sub-trajectories ($n_{H}=10$), $\{(\bar{\bx}_{k:k+h},\bar{\bu}_{k:k+h})\}^{(i,j)}$, each of length $h=7$ are sampled in the region close to the linearisation centre $\zeta^{(i)}=(0,0,0,0)^\T$ (red dot), where $i=1,\dots,n_{a}$ (e.g. in this case $i$ could equal 1, i.e. performing sub-sampling for the first linearisation centre) and $j=1,\dots, n_{H}$ sub-trajectory number.
   	The sub-sampling boundaries are expertly defined for each task (red rectangle).  
   	Sub-trajectories that fall outside of the specified boundaries are not considered (black dots), while those that fall inside are suitable hence sub-sampled (each sub-trajectory is uniquely coloured).
   	The sampled sub-trajectory entries do not repeat and all points of the sub-trajectories are within a bounding box centred at the linearisation centre.
    
    \clearpage
    \section{Learning Curve: Cart-pole}
\begin{figure}[h!]
    	\centering
    	\includegraphics[width=0.75\textwidth]{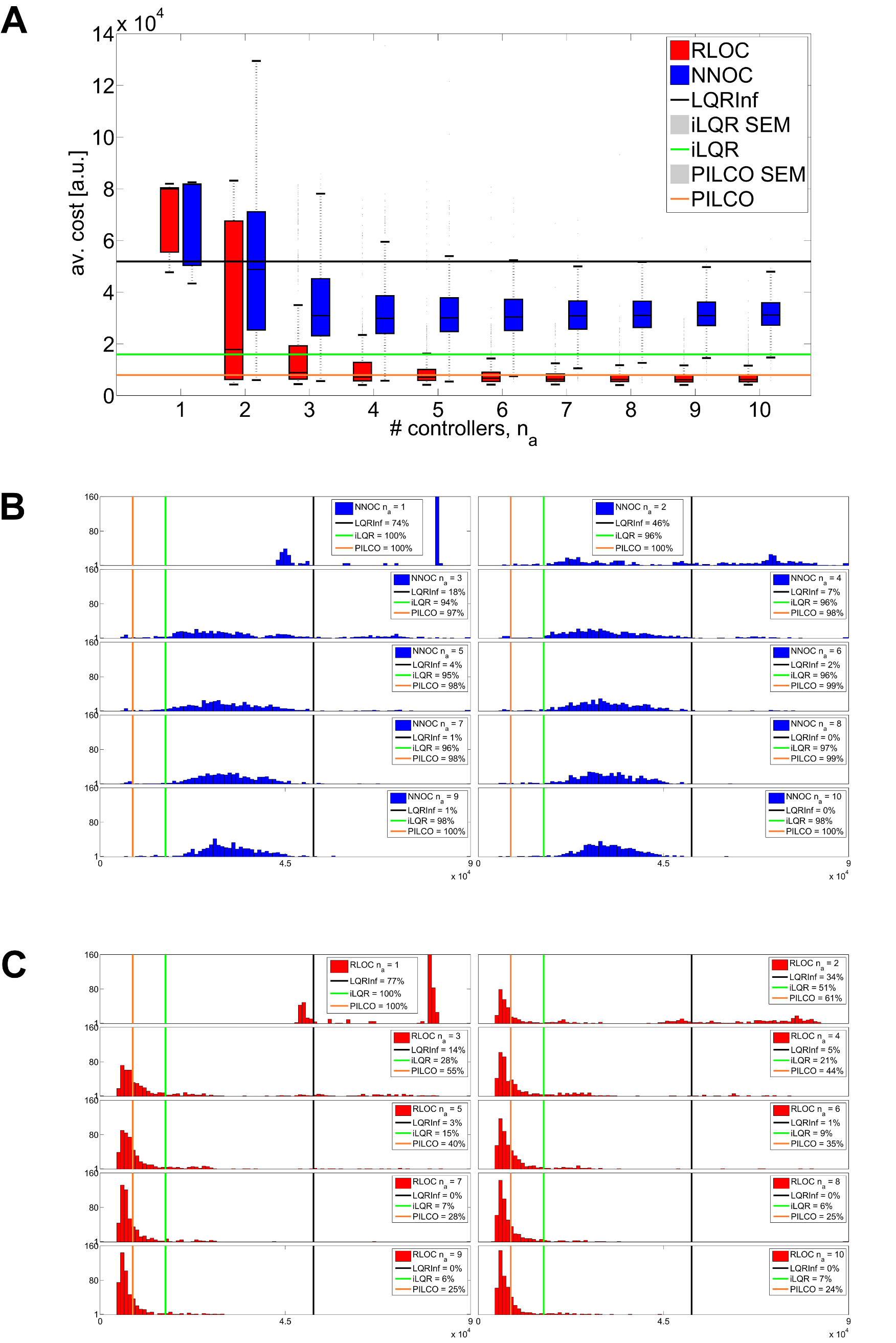}
    	\caption{
    		{\bf Learning curve (cart-pole).} 
    		({\textbf{A}})
    		The performance of RLOC control solution learnt as a function of individual controllers (data points) with respect to engineering state of the art solutions (solid lines), reported using box plot and whiskers diagram, in order to better understand the statistics of the data which was found to be not normally distributed. 
    		({\textbf{B}}) Depicts 10 histograms, with 100 bins each, for the Naive Neighbour Optimal Control (NNOC) average costs when solving the cart-pole task with $a^{(i)},i=1,\dots,10,$ controllers. 
    		({\textbf{C}}) Depicts 10 histograms, with 100 bins each, for the Reinforcement Learning Optimal Control (RLOC) average costs when solving the cart-pole task with $a^{(i)},i=1,\dots,10,$ controllers.
    	}
    	\label{S2_Fig}
    \end{figure}

    Figure \ref{S2_Fig} analyses the average RLOC cost for controlling the cart-pole from 100 equally spaced start positions to the upright target as a function of the number of controllers available to the system, where each controller learning trial is repeated 500 times.
    The average RLOC cost for one learning trial was obtained by taking a mean of the control cost (after learning) across all start configurations of the cart-and-pendulum (100 possibilities).
    The data obtained across trials for each controller was tested for normality using the D'Agostino-Pearson $K^2$ test and was found to not be normally distributed.
    
    Figure \ref{S2_Fig} \textbf{A} depicts the box plot, which allows to observe the following statistics for number of controllers $a^{(i)},i=1,\dots,n{_a}$: lower quartile (q1), median (q2, black line) and upper quartile (q3). 
    The whiskers are calculated using interquartile range (IQR): upper whisker data $\in$ (q3, q3$+$1.5$\times$IQR); lower whisker data $\in$ (q1, q1$-$1.5$\times$IQR); data outside of the whiskers is reported as outliers (black dots). 
    We show that Naive Neighbour Optimal Control (NNOC) cannot outperform iLQR and PILCO, therefore proving the need for high-level learning.
    RLOC's performance for solving the cart-pole task stabilises with the increasing number of controllers and for $n_a\geq$ 7  the interquartile range and the median of the average RLOC costs are below the average cost of PILCO and iLQR (orange and green lines respectively)).
    
    Figure \ref{S2_Fig} \textbf{B} shows that the average costs for the 500 trials of each controller, do not outperform either iLQR or PILCO, with at most being 6$\%$ better. 
    This shows that even though naive switching of controls (to the nearest controller in state space using Euclidean distance) can outperform LQR for solving the cart-pole task, high-level learning of smartly assigning controllers to state space (i.e. RLOC) is necessary.
    
    Figure \ref{S2_Fig} \textbf{C} shows that the average costs for the 500 trials of each controller, gradually start to outperform iLQR and PILCO as the number of controllers increases, where on average for $n_a\geq$7 controllers the average RLOC costs outperform those of iLQR 96$\%$ of the time and PILCO 75$\%$ of the time.

    \clearpage
    \section{Policy Evaluation: Robotic Arm}
    The high-level policy learns the area of state space (i.e. state) to which each controller should be assigned to in order to solve the robotic arm inverse kinematics task.
    
\begin{figure}[h!]
     	\centering
     	\includegraphics[width=1.0\textwidth]{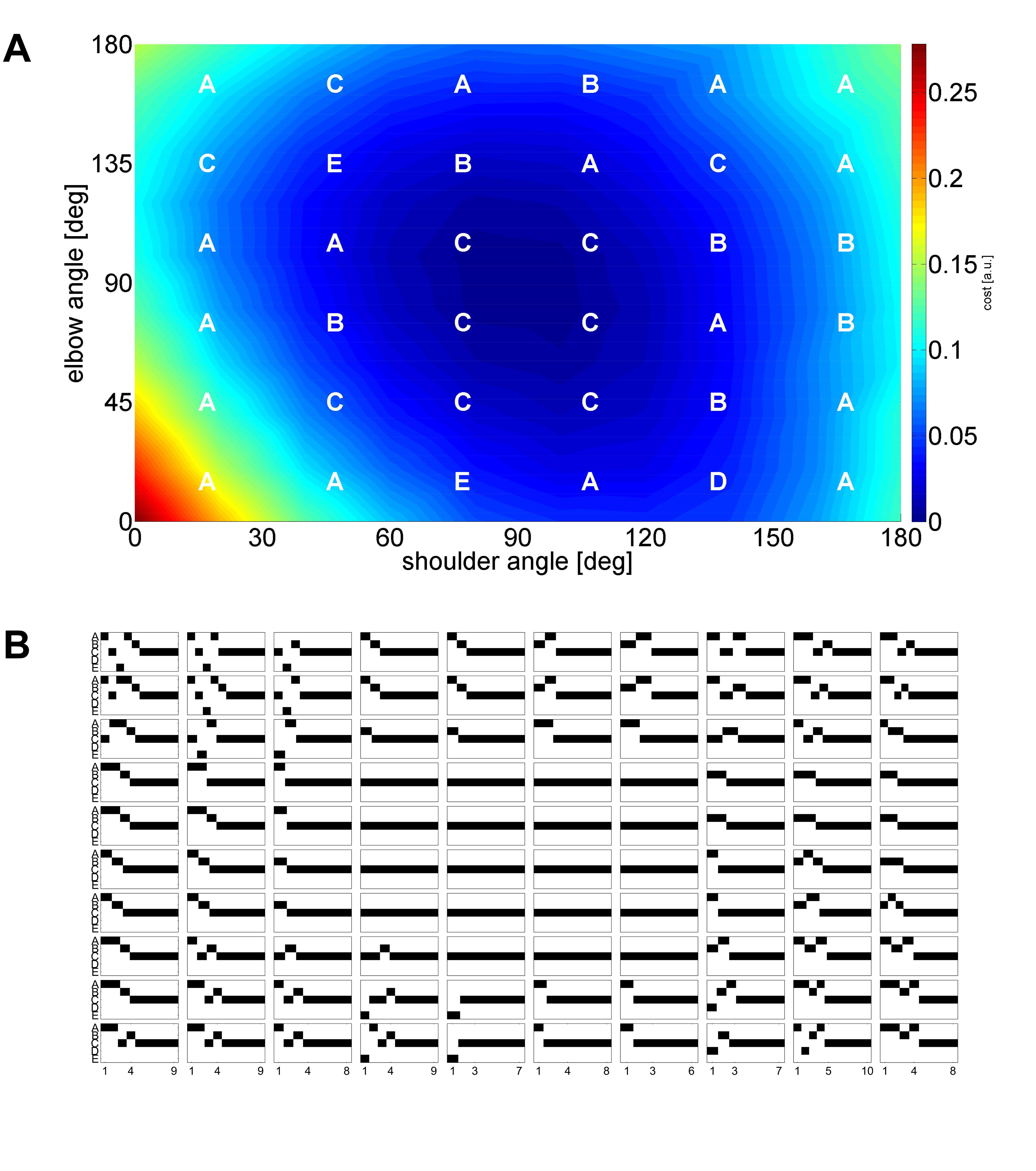}
     	\caption{
     		{\bf Policy evaluation (robotic arm).}
     		({\textbf{A}}) An example robotic arm final policy is evaluated for 5 controllers (labelled A-E) and the cost-to-go function is displayed with superimposed control to state space mapping ($n_{s}= 36$).
     		({\textbf{B}}) Action sequences generated by an RLOC policy during control of the robotic arm to a target from 100 equally spaced initial states and using 5 controllers with $n_K=1000$ steps, $\delta t=0.01$ time step and the same final policy as that used to obtain the value function of (A). 
     	}
     	\label{S3_Fig}
     \end{figure}

    Figure \ref{S3_Fig} \textbf{A} shows the cost-to-go is obtained for $1 \times 10^5$ discretised states and simulation parameters $n_K=1000, \delta t=0.01$. 
    A smooth, convex value function is observed, with the smallest cost incurred for controlling the arm from states near to the target $\bx^\ast = [90\degree,90\degree,0,0]$ and the cost for controlling the arm increasing gradually as the start position moves further and further away from the target. 
    The nonlinearity of the planar robotic arm is mathematically explained by nonlinear dependencies of the elbow angle only (and not the shoulder), which means that it is hardest to move the arm with the elbow fully extended (i.e. elbow angle of $0\degree$) - hence the area of state space in the $(0,0)^\T$ region is the hardest to control, displaying higher cost-to-go (red colour). 
    
    Figure \ref{S3_Fig} \textbf{B} 
    The linearisation centre for each controller of the example policy has the general form $\zeta^{(i)} = [\theta_{1}, \theta_{2}, 0, 0]^\T$, 
    where 
    $\bf{\theta} = 
    \bigl(\begin{smallmatrix} \theta_{1}\\ \theta_{2} \end{smallmatrix} \bigr) =
    \{ 
    \bigl(\begin{smallmatrix} 90\degree\\ 172\degree \end{smallmatrix} \bigr), 
    \bigl(\begin{smallmatrix} 90\degree\\ 97\degree \end{smallmatrix} \bigr),
    \bigl(\begin{smallmatrix} 90\degree\\ 66\degree \end{smallmatrix} \bigr), 
    \bigl(\begin{smallmatrix} 90\degree\\ 13\degree \end{smallmatrix} \bigr),
    \bigl(\begin{smallmatrix} 90\degree\\ 30\degree \end{smallmatrix} \bigr)
    \}$
    for controllers $n_a=1,\dots,5$  respectively.
    Each row within the sub-plot indicates an action (labelled A-E) which, if chosen, is marked with a black dot plotted vs. horizontal axis of the action switch number. 
    The switching of the controls, as dictated by the high-level policy, is allowed once the continuous Actuator State (i.e. that of the optimal control space) $\bx_{k}, k=1,\dots,n_K$, enters a new discrete Symbolic State (i.e. that of the reinforcement learner) $s^{(i)}, i=1,\dots,n_{s}$. 
    The high-level action sequences appear as "activation patterns", which are reused by the RLOC policy when solving the task from areas of Actuator State Space with similar cost-to-go function. 
    Action "C" can be seen to be preferred when brining the arm to the target.  

    \clearpage
    \section{Robotic Arm Dynamics} \label{S1_Appendix}
    Biological systems and their motor control have been extensively studied \cite{li2004iterative,todorov2005generalized} in order to improve the ability of robots to learn from experience rather than being susceptible to repeated execution errors due external calibration references. 
    The human motor system continually specifies muscle lengths and joint angles in order to attain desired arm and hand position (the inverse kinematic problem).
    A specific path trajectory can be achieved through transformation from joint torques to movement (the forward dynamics) \cite{atkeson1989learning} and the pattern of muscle activations to achieve these torques has been shown to be consistent with prediction of torque signals obtained by an optimal control framework.
    
    The arm is modelled as two weightless rods, with masses centred near the middle of each link with friction acting on the joint. 
    The simulated robot arm is a function of joint positions ($\theta_{1}$ shoulder angle and $\theta_{2}$ elbow angle), joint velocities ($\dot{\theta}_{1}, \dot{\theta}_{2}$) and joint accelerations ($\ddot{\theta}_{1}, \ddot{\theta}_{2}$).
    Here, the robotic arm (RA) task consists of starting from any state $\bx_{0} = [\theta_{1}, \theta_{2}, \dot{\theta_{1}},\dot{\theta_{2}}]^\T$, where $\theta_{1}\in[0,\pi]$ is the shoulder angle, $\theta_{2}\in[0,\pi]$ is the elbow angle, $\dot{\theta_{1}}\in[{-300\degree},{300\degree}]$ is the shoulder angular velocity, $\dot{\theta_{2}}\in[{-300\degree},{300\degree}]$ is the elbow angular velocity and controlling the robotic arm to the target state of $\bx^{\ast} = [90\degree,90\degree,0,0]^\T$. 
    The Actuator State vector $\bx_{k}$ at time step $k$, the Actuator State dynamics vector $\dot{\bx}_{k}$ and the forward dynamics equation are given by
    \[
    \bx_{k}
    = (\theta_1,\theta_2,\dot{\theta}_1,\dot{\theta}_2)^\T
    \qquad
    \dot{\bx}_{k}
    = (\dot{\theta}_1,\dot{\theta}_2,\ddot{\theta}_1,\ddot{\theta}_2)^\T
    \qquad
    \mathbf{\ddot{\theta}}
    = \mathcal{M}(\mathbf{\theta})^{-1}(\mathbf{\tau} - \mathcal{C}(\mathbf{\theta},\mathbf{\dot{\theta}}) - \mathcal{B}\mathbf{\dot{\theta}})
    \]
    where $\mathbf{\theta}\in[0,180\degree]^2$ is the joint angle vector, $\mathcal{M}(\mathbf{\theta})$ is a positive semi-definite inertia matrix, $\mathcal{C}(\mathbf{\theta}, \mathbf{\dot{\theta}})\in\mathbb{R}^2$ is a vector of Centripetal and Coriolis forces, $\mathcal{B}\in\mathbb{R}^{2\times2}$ is the joint friction matrix and $\mathbf{\tau}\in[-10,10]^2$ is the joint torque (defined to be the control $\mathbf{u} = \mathbf{\tau}$) \cite{li2007}. 
    
    We simulated this robotic arm system using the Runge-Kutta approximation method, which approximated the solutions to ordinary differential equations of the form Eq. \ref{eq:xdot}, where $\bx_{k+1}=\bx_{k} + \frac{1}{6}(l_1 + 2l_2 + 2l_3 + l_4) + \mathcal{O}(n^5), 
    ~l_1=\delta t\mathbf{f}(\bx_{k},\bu_{k}),
    ~l_2=\delta t\mathbf{f}(\bx_{k}+\frac{1}{2}l1,\bu_{k}),
    ~l_3=\delta t\mathbf{f}(\bx_{k}+\frac{1}{2}l2,\bu_{k}),
    ~l_4=\delta t\mathbf{f}(\bx_{k}+l3,\bu_{k})$
    and used the state and control penalties of $W$ = diag(30, 30, 0, 0), $Z$ = diag(1,1) (see Table \ref{Table3} for full list of parameters used).
    \begin{table}[!ht]
    	\centering
    	\fontsize{10pt}{10pt}\selectfont
    	\caption{\bf{Robotic Arm Simulation Parameters}}
    	{\begin{tabular}{|l|l|l|l|}
    			\hline
    			\bfseries Parameter Description & \bfseries Symbol & \bfseries Value & \bfseries SI Unit\\ \hline
    			Length of link 1 & $l_{1}$ & 0.3 & [$m$]\\ \hline
    			Length of link 2 & $l_{2}$ & 0.33 & [$m$]\\ \hline
    			Mass of link 1 & $m_{1}$ & 1.4 & [$kg$]\\ \hline
    			Mass of link 2 & $m_{2}$ & 2.5 & [$kg$]\\ \hline
    			Centre of mass for link 1 & $C_{1}$ & 0.11 & [$m$]\\ \hline
    			Centre of mass for link 2 & $C_{2}$ & 0.165 & [$m$]\\ \hline
    			Moment of inertia for link 1 & $i_{1}$ & 0.025 & [$kg~m^2$]\\ \hline   
    			Moment of inertia for link 2 & $i_{2}$ & 0.072 & [$kg~m^2$]\\ \hline  
    			Joint friction & $B_j$ & $\bigl(\begin{smallmatrix} 0.5 ~0.1\\ 0.1 ~0.5 \end{smallmatrix} \bigr)$ & [$N$]\\ \hline 
    			Maximum joint torque & $u_{max}$ & 10 & [$N~m$]\\ \hline 
    			Maximum negative torque & $u_{min}$ & -10 & [$N~m$]\\ \hline
    			Dimensionality of state vector & $l$ & 4 & [$a.u.$]\\ \hline
    			Dimensionality of control vector & $m$ & 2 & [$a.u.$]\\ \hline
    			Number of naive start states & $n_{\widetilde{s}}$ & 790 & [$a.u.$]\\ \hline
    			LDSi sub-trajectory length & $h$ & 7 & [$a.u.$]\\ \hline
    			LDSi average number of sub-trajectories & $n_H$ & 6500 & [$a.u.$]\\ \hline
    			Number of actuator space simulated steps & $n_{K}$ & 300 & [$a.u.$]\\ \hline
    			Low-level control simulation time step & $\delta t$ & 0.01 & [$sec$]\\ \hline  
    			Target state & ${\bx}^{\ast}$ & [$90,90,0,0$] & [$deg,deg,deg/s,deg/s$]\\\hline
    			Number of discretised dimensions of $\bx$ & $n_{l}$ & 2 & [$a.u.$]\\ \hline
    			Number of Symbolic States & $n_{s}$ & 36 & [$a.u.$]\\ \hline  
    	\end{tabular}}
    	\begin{flushleft}
    		Parameters used to simulate RA task, learn local linear models with LDSi and define the target state.
    	\end{flushleft}
    	\label{Table3}
    \end{table}

    \section{Cart-Pole Dynamics} \label{S2_Appendix}
    The cart-pole (CP)system is a standard, but challenging, benchmark task in modern control theory.
    The task consists of actively balancing the pole in the upright position (unstable equilibrium), with the cart stopping at a pre-defined position on the track, while control is limited to forces applied laterally to the cart along the horizontal plane.
    Due to the fact that the rotation of the pole cannot be controlled directly, the cart-pole is an underactuated system, this makes standard nonlinear control techniques ineffective \cite{Lozano2000}.  
    RLOC attempts to learn the control of the cart-pole system from all positions in the available state space and is not given access to the dynamics a priori, but has to learn necessary dynamics online.
    
    The simulated cart-pole is a function of cart position $z$, cart velocity $\dot{z}$, pendulum angle $\theta$ and pendulum angular velocity $\dot{\theta}$, with friction applied between the cart and the track.
    Here, the cart-pole task consists of starting from any state $\bx_{0} = [z, \dot{z}, \theta, \dot{\theta}]^\T$, where $z\in[0]$ is the position of the cart, $\dot{z}\in[0]$ is the velocity of the cart, $\theta\in[-180\degree,180\degree]$ is the pendulum angle, $\dot{\theta}\in[-250\degree,250\degree]$ is the velocity of the pendulum (with ${\theta} = 0\degree$ defined to be the upright, unstable equilibrium) and controlling the cart-pole to the target state of $\widetilde{\bx}^{\ast} = \mathbf{0}$.
    The Symbolic State vector $\bx_{k}$ at time step $k$, the Symbolic State dynamics vector $\dot{\bx}_{k}$ and the inverse kinematics are given by
    \[
    \bx_{k} = (z,\dot{z},\theta,\dot{\theta})^\T
    \qquad
    \dot{\bx}_{k} = (\dot{z},\ddot{z},\dot{\theta},\ddot{\theta})^\T
    \qquad
    {M}(\mathbf{q}){\ddot{\mathbf{q}}}^2 + {C}(\mathbf{q},{\dot{\mathbf{q}}}){\dot{\mathbf{q}}} + {G}(\mathbf{q}) = \mathbf{\tau}
    \]
    where 
    \[
    \begin{array}{c@{\quad}c}
    
    {M}(\mathbf{q}) = 
    \left(
    \begin{array}{ccc}
    {{m}_p + {m}_c} & {m}_p{l}\cos{\theta}\\
    {m}_p{l}\cos{\theta} & {m}_p{l}^2\\
    \end{array}
    \right),
    &
    {C}(\mathbf{q},{\dot{\mathbf{q}}}) = 
    \left(
    \begin{array}{ccc}
    0 & -{m}_p{l}{\dot{\theta}}\sin{\theta}\\
    0 & 0\\
    \end{array}
    \right)
    \end{array}
    \]
    and ${\mathbf{q}} = [z,\theta]^\T$, $\mathbf{\tau} = [{F},0]^\T$
    
    To simulate the cart-pole system, we again used the Runge-Kutta approximation method and used the state and control penalties of $W$ = diag(30, 3, 2000, 200), $Z$ = (1) (see Table \ref{Table4} for full list of parameters used).
    \begin{table}[!ht]
    	\centering
    	\fontsize{10pt}{10pt}\selectfont
    	\caption{\bf{Cart Pole Simulation Parameters}}
    	{\begin{tabular}{|l|l|l|l|}
    			\hline
    			\bfseries Parameter Description & \bfseries Symbol & \bfseries Value & \bfseries SI Unit\\ \hline
    			Length of pendulum & $l$ & 0.6 & [$m$]\\ \hline
    			Mass of pendulum & $m_{p}$ & 0.5 & [$kg$]\\ \hline
    			Mass of cart & $m_{c}$ & 0.5 & [$kg$]\\ \hline
    			Acceleration due to gravity & $g$ & 9.80665 & [$m/{s^2}$]\\ \hline
    			Pendulum friction constant & $B_{p}$ & 0 & [$N/m/s$]\\ \hline
    			Cart friction constant & $B_{c}$ & 0.1 & [$N/m/s$]\\ \hline  
    			Maximum force & $u_{max}$ & 20 & [$N$]\\ \hline
    			Maximum negative force & $u_{min}$ & -20 & [$N$]\\ \hline
    			Number of naive start states & $n_{\widetilde{s}}$ & 253 & [$a.u.$]\\ \hline
    			LDSi sub-trajectory length & $h$ & 20 & [$a.u.$]\\ \hline
    			LDSi average number of sub-trajectories & $n_H$ & 170 & [$a.u.$]\\ \hline
    			Number of actuator space simulated steps & $n_{K}$ & 300 & [$a.u.$]\\ \hline
    			Low-level control simulation time step & $\delta t$ & 0.01 & [$sec$]\\ \hline
    			Target state & $\widetilde{\bx}^{\ast}$ & \bf{0} & [$m,m/s,deg,deg/s$]\\ \hline
    			Number of discretised dimensions of $\bx$ & $n_{l}$ & 2 & [$a.u.$]\\ \hline
    			Number of Symbolic States & $n_{s}$ & 49 & [$a.u.$]\\ \hline
    	\end{tabular}}
    	\begin{flushleft}
    		Parameters used to simulate CP task, learn local linear models with LDSi and define the target state.
    	\end{flushleft}
    	\label{Table4}
    \end{table}

\end{document}